\documentclass[preprint,authoryear,3p]{elsarticle}
\usepackage{amssymb,amsthm,amsmath}
\usepackage{graphicx}
\usepackage{subfig}
\usepackage{float}
\usepackage{multirow}
\usepackage{longtable,lscape}
\usepackage{cases}
\usepackage{url}
\usepackage{caption}
\usepackage{color}
\usepackage{amsthm}
\usepackage{algpseudocode,algorithm,algorithmicx}
\usepackage[pagewise]{lineno}

\usepackage{enumitem}
\DeclareMathOperator*{\argmax}{arg\,max}  % in your preamble
\usepackage{fancyhdr}
\usepackage{fullpage}

\theoremstyle {plain}% default

\newcounter{x}

\theoremstyle{definition}
\newtheorem{defn}{Definition}[section]

\theoremstyle{remark}
\newtheorem*{rem}{Remark}

\usepackage{titlesec}
\titleformat{\paragraph}
{\normalfont\normalsize\bfseries}{\theparagraph}{1em}{}
\titlespacing*{\paragraph}
{0pt}{3.25ex plus 1ex minus .2ex}{1.5ex plus .2ex}

\usepackage{color}

\makeatletter
\def\ps@pprintTitle{%
	\let\@oddhead\@empty
	\let\@evenhead\@empty
	\def\@oddfoot{\reset@font\hfil\thepage\hfil}
	\let\@evenfoot\@oddfoot
}
\makeatother

\numberwithin{equation}{section}

\begin{document}

%\linenumbers

\begin{frontmatter} 
		
		
		
%\title{Where to Find Next Passengers on E-hailing Platforms? - A Markov Decision Process Approach}

\title{Optimal Passenger-Seeking Policies on E-hailing Platforms Using Markov Decision Process and Imitation Learning}

\noindent \textcolor{blue}{Published in: Transportation Research Part C 111 (2020) 91-113.}

\noindent \textcolor{blue}{Please cite this Paper as: Shou, Z., Di, X., Ye, J., Zhu, H., Zhang, H., Hampshire, R., 2020. Optimal passenger-seeking policies on E-hailing platforms using Markov decision process and imitation learning. Transportation Research Part C: Emerging Technologies 111, 91–113. DOI: 10.1016/j.trc.2019.12.005}
		
		% \date{\today}
		
		
\author[cu]{Zhenyu Shou}

%\author[cu]{Zhaobin Mo}

\author[cu,dsi]{Xuan Di\corref{cor}}
\ead{sharon.di@columbia.edu}

\author[didi]{Jieping Ye}

\author[didi]{Hongtu Zhu}

\author[tongji]{Hua Zhang}

\author[ford,UMTRIaddr]{Robert Hampshire}

\cortext[cor]{Corresponding author. Tel.: +1 212 853 0435;}

\address[cu]{Department of Civil Engineering and Engineering Mechanics, Columbia University}
\address[dsi]{Data Science Institute, Columbia University}
\address[didi]{Didi Chuxing Inc., Beijing, China}
\address[tongji]{National Maglev Transportation Engineering R\&D Center, Tongji University, Shanghai, China}
\address[UMTRIaddr]{University of Michigan Transportation Research Institute, University of Michigan, Ann Arbor} 
\address[ford]{Ford School of Public Policy, University of Michigan, Ann Arbor}

\begin{abstract}
	
Vacant taxi drivers' passenger seeking process in a road network generates additional vehicle miles traveled, adding congestion and pollution into the road network and the environment. 
This paper aims to employ a Markov Decision Process (MDP) to model idle e-hailing drivers' optimal sequential decisions in passenger-seeking. 
%While there exist a few studies that applied MDPs to taxi drivers' searching behavior, these studies were primarily focused on modeling traditional taxi drivers' behavior. 
Transportation network companies (TNC) or e-hailing (e.g., Didi, Uber) drivers exhibit different behaviors from traditional taxi drivers because e-hailing drivers do not need to actually search for passengers. Instead, they reposition themselves so that the matching platform can match a passenger. 
Accordingly, we incorporate e-hailing drivers' new features into our MDP model. The reward function used in the MDP model is uncovered by leveraging an inverse reinforcement learning technique.
We then use 44,160 Didi drivers' 3-day trajectories to train the model. 
To validate the effectiveness of the model, a Monte Carlo simulation is conducted to simulate the performance of drivers under the guidance of the optimal policy, which is then compared with the performance of drivers following one baseline heuristic, namely, the local hotspot strategy. 
The results show that our model is able to achieve a 17.5\% improvement over the local hotspot strategy in terms of the rate of return. The proposed MDP model captures the supply-demand ratio considering the fact that the number of drivers in this study is sufficiently large and thus the number of unmatched orders is assumed to be negligible. To better incorporate the competition among multiple drivers into the model, we have also devised and calibrated a dynamic adjustment strategy of the order matching probability. 
% in the morning peak, 10.67\% during the off-peak, and 13.20\% in the evening peak and improve the utilization rate of the vehicle. %The result also shows that the proposed model works the best for the evening peak when the performance of the real drivers is relatively worse than that in other two time intervals.

\end{abstract}
		
		
\end{frontmatter}
	
\section{Motivation}

	Taxi, complementary to massive transit systems such as bus and subway, provides flexible-route door-to-door mobility service. However, taxi drivers usually have to spend 35-60 percent of their time on cruising to find the next potential passenger \citep{powell_towards_2011}. Such passenger-seeking process not only decreases taxi drivers' income but also generates additional vehicle miles traveled, adding congestion and pollution into the increasingly saturated roads. 

	Cruising is primarily caused by an imbalance between travel demand and supply. Market regulation \citep{yang_demandsupply_2002} or taxi fare structure design \citep{yang_nonlinear_2010,he_pricing_2018,battifarano_predicting_2019} were proposed respectively to balance taxi travel demand and supply. A network equilibrium model was developed \citep{yang_network_1998, wong_network_1998} to capture the spatial imbalance between travel demand and supply, where a logit-based probability was introduced to describe the meeting between a vacant taxi and a waiting passenger. This model, in which a taxi driver is supposed to minimize the individual search time for the next passenger, is further extended to incorporate congestion effects and customer demand elasticity \citep{wong_modeling_2001}, to include the fare structure and fleet size regulation \citep{yang_demandsupply_2002}, to consider multiple user classes, multiple taxi modes, and customer hierarchical modal choice \citep{wong_modeling_2008}, and to use a meeting function to describe the search frictions between vacant taxis and waiting passengers \citep{yang_equilibria_2010, yang_equilibrium_2011, yang_taxi_2014}. Recently, \cite{di_unified_2019} proposed a unified equilibrium framework to model the shared mobility in congested road network.

	As taxis GPS trajectories become increasingly available, qualitative analysis has been performed to uncover drivers' actual searching strategy. \cite{liu_uncovering_2010} found that drivers with higher profits prefer to choose routes with higher speed in both operational and idle states. \cite{li_hunting_2011} discovered that hunting is a more efficient strategy than waiting by comparing profitable and non-profitable drivers. Several logit-based quantitative models were developed to capture idle drivers' searching behavior \citep{szeto_time-dependent_2013, sirisoma_empirical_2010, wong_cell-based_2014, wong_modelling_2014, wong_sequential_2015, wong_two-stage_2015}. The bilateral searching behavior (i.e., taxi searching for customers and customers searching for taxis) was modeled through an absorbing Markov chain approach \citep{wong_modeling_2005}. A probabilistic dynamic programming routing model was proposed to capture the taxi driver's routing decisions at intersections \citep{hu_modeling_2012}. Furthermore, a two-layer approach, in which the first layer models the driver's pick-up location choice and the second layer accounts for the driver's detailed route choice behavior, was presented \citep{tang2016two}. Recently, \cite{zhang_hunting_2019} proposed an image-based representation of taxi drivers' passenger searching strategies and identified twenty four  strategies using a dataset collected in Shenzhen, China.

	Upon the understanding of drivers' searching behavior, recommendations can be provided to idle drivers on where to find the next passenger. An accurate prediction of both taxi supply \citep{phithakkitnukoon_taxi-aware_2010} and demand \citep{moreira-matias_predictive_2012, markou_predicting_2019,wang_understanding_2019,alemi_what_2019} as well as travel time \citep{Tan2018WhenWY, zhang_novel_2019} are stepping stones to these recommendations. The objectives that recommendations aim to achieve include the minimization of waiting time at the recommended location \citep{hwang_effective_2015} or of the distance between the current location and the recommended location \citep{powell_towards_2011, hwang_effective_2015}, and the maximization of the expected fare for the next trip \citep{powell_towards_2011, hwang_effective_2015}, of the probability of finding a passenger \citep{ge_energy-efficient_2010}, or of the potential profit of a driver \citep{qu_cost-effective_2014, yuan_where_2011}.  

	The aforementioned studies mainly focused on the recommendation of the cruising routes or next cruising locations at the immediate next step without considering the optimization of long-run payoffs. A recommended customer searching strategy may help a driver to get an order as fast as possible but may not maximize this driver's overall profit in one day. Models which can capture drivers' long-term optimization strategy are needed. In recent years, Markov Decision Process (MDP) becomes increasingly popular in optimizing a single agent's sequential decision-making process given a period of time \citep{puterman_markov_1994}. Several studies \citep{rong_rich_2016, zhou_optimizing_2018, verma_augmenting_2017, gao_optimize_2018, yu_markov_2019} have employed MDPs to model idle drivers' optimal searching strategy. In an MDP, an idle driver is an agent who makes sequential decisions of where to go to find the next passenger in a stochastic environment. The environment is characterized by a Markov process and transitions from one state to another once an action is specified by the idle driver. The driver aims to select an optimal policy which optimizes her long-term expected reward. Dynamic programming or Q-learning approaches are commonly used to solve an MDP \citep{sutton_introduction_1998}. Table~(\ref{tab:mdp}) summarizes the existing studies using MDPs for passenger-seeking optimization. Note that in e-hailing, there is actually no passenger seeking because it is the e-hailing paltform that matches an idle e-hailing driver to a passenger. However, e-hailing drivers still need to \emph{reposition} themselves in order to get better chance of getting matched to a passenger request. In this paper, we will use the terminologies passenger seeking and repositioning interchangably.
\begin{table}[H]
	\centering\caption{Existing MDP based models on passenger-seeking strategy}
	\label{tab:mdp}
	\begin{tabular}{ p{40pt}| p{60pt}| p{60pt}| p{100pt}| p{70pt} | p{50pt}} 
		\hline
		Reference & Network representation & State space & Action space & Reward & Algorithm \\ \hline
		\cite{rong_rich_2016} & Grid world & (grid id, time, incoming direction) & moving to a neighboring grid or staying in the current grid & Taxi fare & Dynamic programming\\ \hline 
		\cite{verma_augmenting_2017} & Grid world (static and dynamic zone structure) & (day-of-week, grid id, time-interval) & moving to any chosen grid (proposed an action detection algorithm) & Taxi fare - traveling distance cost - time cost & Q-learning (Monte Carlo) \\ \hline
		\cite{gao_optimize_2018} & Grid world & (grid id, operating status) & driving vacantly to neighboring grids to search, finding a passenger in the current grid, waiting static at the same spot & the ratio of the occupied taxi trip mileage to the previous empty mileage & Q-learning (Temporal Difference) \\ \hline
		\cite{yu_markov_2019} & Link node & (node id, indicator of the current pick-up drop-off cycle) & outgoing links from the current node & taxi fare - operating cost & Value iteration \\ \hline
		\cite{lin_efficient_2018} & Grid world & (grid id, time interval, global state) & moving into a neighboring grid or staying in the current grid & taxi fare - operating cost & Reinforcement Learning (Deep Q learning) \\ \hline
	\end{tabular}
\end{table}

\begin{figure}[H]
	\centering 
	\subfloat[Traditional taxi driver]{\includegraphics[scale=.5]{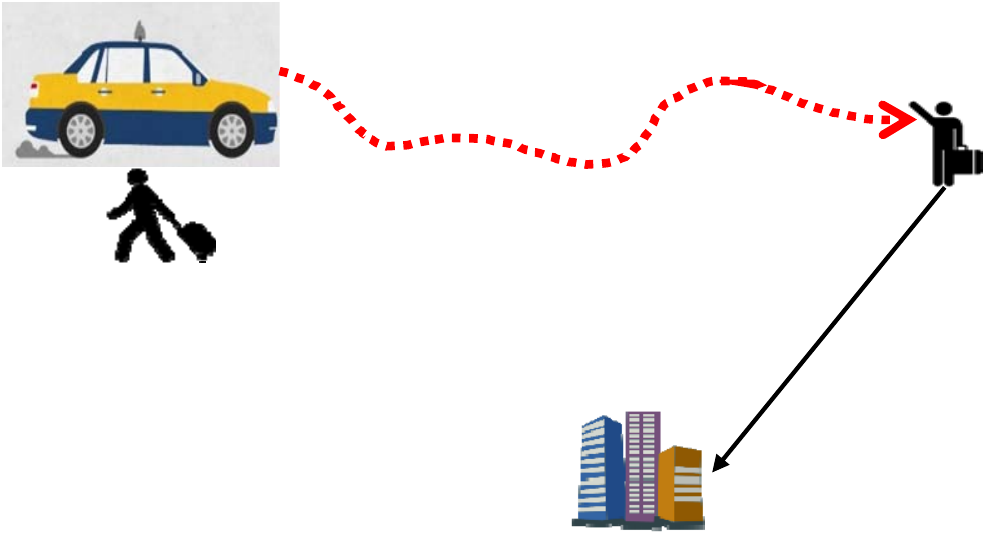}\label{subfig:taxi}}~~~
	\subfloat[E-hailing driver: no searching]{\includegraphics[scale=.5]{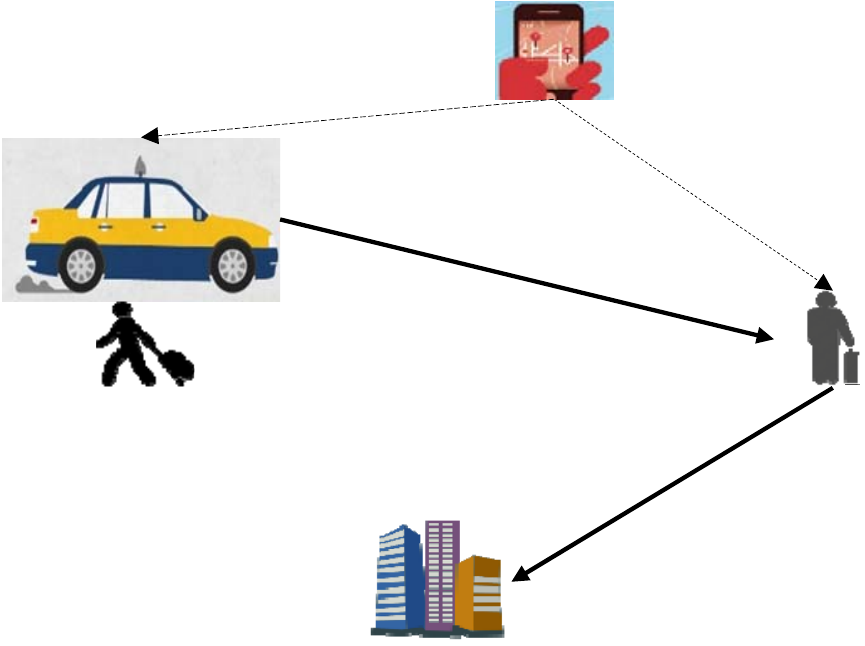}\label{subfig:e-hail_nosearch}}~~~
	\subfloat[E-hailing driver: picking up in a different location]{\includegraphics[scale=.5]{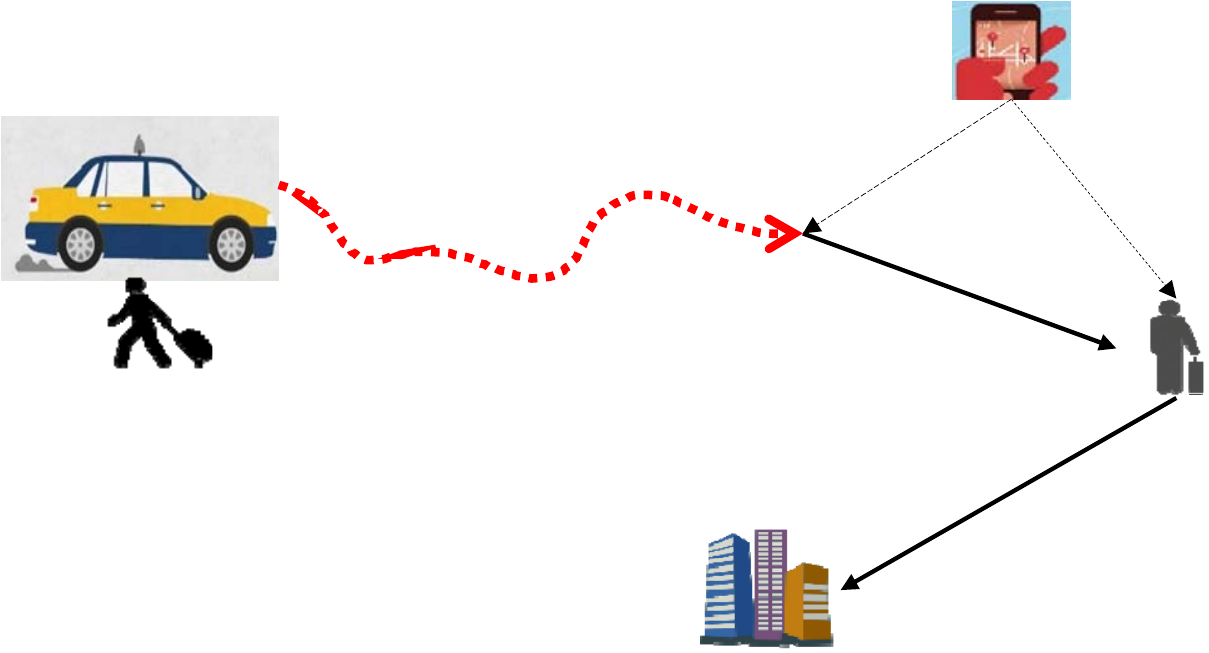}\label{subfig:e-hail_match}}
	\caption{Sequential decision-making processes for traditional and e-hailing taxi drivers}
	\label{fig:e-hail_taxi} 
\end{figure}

The existing MDP models were primarily developed for traditional taxi drivers' sequential decision-making where a driver has to see a passenger before a match happens. In other words, an idle driver's searching process ends only when this driver sees a passenger and the passenger accepts the ride (see Figure~(\ref{subfig:taxi})). 
E-hailing applications (such as Didi and Uber), on the other hand, offer an online platform to match a driver with a passenger even when they are not present in the same space at the same time \citep{he_modeling_2015, qian_taxi_2017}.  
%Accordingly, e-hailing drivers do not have passenger-seeking behavior. 
In other words, even when an idle driver sees a passenger waiting on the roadside, as long as the e-hailing platform does not match them, the driver cannot give a ride to the passenger. 
However, it does not mean e-hailing drivers always stay at the previous drop-off spot and wait for the platform to match. 
Drivers tend to \emph{reposition} themselves so that the platform can find them a match sooner. 
As a result, the decision-making process of e-hailing drivers is quite different from the traditional taxi drivers in the following aspects:
\begin{enumerate}
	\item An e-hailing driver may receive a matched order before she drops off the previous passenger, thus there is no passenger seeking (see Figure~(\ref{subfig:e-hail_nosearch})).
	\item Different from traditional taxi that a driver has to see a passenger to find a match, e-hailing platforms very likely find a match even when the driver and the passenger are spatially far from each other. In other words, a driver's search process may end before a passenger is picked up (see Figure~(\ref{subfig:e-hail_match})).
\end{enumerate}

Because of the inherent differences in drivers' decision-making, this paper aims to develop an MDP to model e-hailing drivers' sequential decision-making in searching for the next passenger. 
44,160 Didi drivers' 3-day GPS trajectories are used to calibrate and validate our model. Previously, there is research using Didi's data for the study of large-scale fleet management \citep{lin_efficient_2018} and large-scale order dispatch \citep{xu_large-scale_2018} in e-hailing platforms.

The major contributions of this paper are as follows: %(1) The proposed MDP model incorporates e-hailing drivers' new characteristic features. The derived optimal policy shows that with the help of an e-hailing platform, drivers can simply choose to wait or stay within the current grid, especially when they are in the busy city area, which is e-hailing specific. 
(1) In stead of following the literature where a known reward function is given based on some prior knowledge or assumptions \citep{rong_rich_2016,verma_augmenting_2017, lin_efficient_2018, yu_markov_2019}, this work unveils the underlying reward function of the overall e-hailing driver population and crafts a novel reward function which explains the behaviors of drivers with a relatively small radius of gyration and thus paves the way for future research on discovering the underlying reward mechanism in a complex and dynamic e-hailing market. With the incomplete and noisy observed policy, this work first extracts the underlying reward function and then solves an MDP to derive the optimal policy which completes and corrects the observed policy.
(2) To the best of our knowledge, this is the first study using large amounts data to devise and calibrate a dynamic adjustment strategy of the order matching probability to address the competition among multiple drivers. The strategy essentially attenuates the order matching probability in an exponential manner for subsequent drivers to be guided into a grid when some drivers have already entered the grid. The strategy is further verified to be efficient in providing different recommendations for multiple drivers.  %(3) The derived policy is validated to be able to guide drivers into grids with a relatively high order matching probability and on average better passenger requests (i.e., requests with a higher profit per unit time). For an agent, a shorter request is not necessarily more preferable. In other words, it is the profit per unit time instead of the duration or distance that determines the goodness of a passenger request. } %This explains the reason for the outperformance of the agent following the optimal policy over real drivers, top 10\% drivers, and the agent following the local hotspot heuristic.}

The remainder of the paper is organized as follows. Section~\ref{sec:mdp} introduces our modified MDP model and details definitions of states, actions, and state transitions and the process of extracting parameters from the data. Section~\ref{sec:multi_mdp} presents the proposed dynamic adjustment strategy of the order matching probability and details the calibration process. Section~\ref{sec:data} introduces the data we used in this research and presents the results, including the derived optimal policy and the Monte Carlo simulation. Section~\ref{sec:conclusion} concludes the paper and provides some future research directions.

\section{Markov Decision Process (MDP) for a single agent} \label{sec:mdp}
\subsection{Preliminaries}

An MDP is specified by a tuple $(S, A, R, P, s_0)$, where $S$ denotes the state space, $A$ stands for the allowable actions, $R$ collects rewards, $P$ defines a state transition matrix, and $s_0$ is the starting state. Given a state $s_t = s \in S$ and a specified action $a_t = a \in A$ at time $t$, the probability of reaching state $s'$ at time $t+1$ is determined by the probability transition matrix $P(s,a,s')$, which is defined as
\begin{equation}
	P(s,a,s')\equiv Pr(s_{t+1}=s'|s_t = s, a_t = a)
	\label{eqn:transition}
\end{equation}

From the initial state $s_0$, the process proceeds repeatedly by following the dynamics of the environment defined by the Equation~(\ref{eqn:transition}) until a terminal state (i.e., either the current time exceeds the terminal time or the current state is an absorbing state) is reached.
An MDP satisfies the Markov property which essentially says that the future process is independent on the past given the present, i.e.,
\begin{equation}
	Pr(s_{t+1}=s'|s_{0}, a_{0}, \cdots, s_{t-1}, a_{t-1}, s_t, a_t) = Pr(s_{t+1}=s'|s_t=s, a_t=a).
\end{equation}

There are two types of value functions in MDPs, namely, a state value $V(s)$ and a state-action value $Q(s,a)$. 
%The value function of a state $s$ (or a state-action pair) is an expectation of all future rewards the agent can get from $s$. %Obviously, the future rewards that an agent can expect are dependent on the actions that the agent will take. 
The actions that an agent will take form a policy $\pi$, which is a mapping from a state $s$ and an action $a$ to the probability $\pi(a|s)$ of taking action $a$ at state $s$. %Accordingly, the value function is dependent on the policy $\pi$. 
Then the value function of a state $s$ by following the policy $\pi$, denoted as $V_{\pi}(s)$, can be taken as the expectation of the future rewards, i.e., 
\begin{equation}
	V_{\pi}(s)=\mathbb{E}_{\pi}\left[\sum_{k=0}^{K}\gamma^k r_{t+k+1}|s_t=s\right]
\end{equation}
where $\gamma$ is a discount factor. The state-action value of taking action $a$ at state $s$ by following policy $\pi$ is
\begin{equation}
	Q_{\pi}(s,a)=\mathbb{E}_{\pi}\left[\sum_{k=0}^{K}\gamma^k r_{t+k+1}|s_t=s, a_t=a\right]
\end{equation}

The value function $V_{\pi}(s)$ is actually a weighted average of the state-action value $Q_{\pi}(s, a)$, i.e., 
\begin{equation}
	V_{\pi}(s) = \sum_{j = 1}^{J} \pi(a = a_j|s) Q_{\pi}(s, a = a_j) 
	\label{eqn:value}
\end{equation}
where $\pi(a = a_j|s)$ is again the probability of taking action $a_j$ at state $s$ according to policy $\pi$, and $J$ is the total number of actions that are allowed to be taken in state $s$.

Several algorithms have been developed to solve the MDP, i.e., to derive the optimal policy, and the corresponding value functions, such as the dynamic programming method and the Q-learning approach \citep{sutton_introduction_1998}. The dynamic programming algorithm is used in this work and will be explained later in Section~(\ref{sec:objective}). %The Q-learning approach basically says that state-action value is supposed to be updated iteratively as
%\begin{equation}
%Q(s,a) \leftarrow (1-\eta) Q(s,a) + \eta (r(s,a)+\gamma max_{a'}Q(s',a')-Q(s,a)) 
%\end{equation}
%where $\eta$ is the learning rate and $r$ is the instant reward the agent can collect by specifying action $a$ at state $s$. 
Furthermore, these two types of value functions at optimality are related by the following mechanism
\begin{equation}
 	V(s) = max_a Q(s,a) 
\end{equation}
The rationale underlying the relationship at optimality is simply to choose a policy which maximizes the value function expressed in Equation~(\ref{eqn:value}). For example, when an agent is at state $s$, the optimal policy simply suggests the agent to take an action $a$ with the largest state-action value, i.e., $\pi(\argmax_a{Q(s,a)}|s) = 1$ (i.e., the probability of taking action $\argmax_a{Q(s,a)}$ in state $s$ is 1). Accordingly, the state-action value at optimality can be written as
\begin{equation}
	Q(s,a) = \sum_{s'} Pr(s'|s,a) (V(s') + r(s,a,s'))
	\label{eqn:v_and_q}
\end{equation}
where $Pr(s'|s,a)$ is the probability of landing in state $s'$ after taking action $a$ in state $s$, and $r(s,a,s')$ is the reward for choosing action $a$ at state $s$ and landing in state $s'$. 

\subsection{MDP for e-hailing drivers}

In this section, we will develop an MDP model for e-hailing drivers' stochastic passenger seeking process. Notations which will be used in the subsequent analysis are listed in Table~(\ref{tab:parameters}).

\begin{table}[H]
	\centering\caption{Notations}
	\label{tab:parameters}
	\begin{tabular}{| p{80pt}| p{280pt}| }
		\hline
		Variable & Explanation  \\ \hline
		$l$ & Index of the current grid \\ \hline
		$t$ & Current time \\ 	\hline
		$I$ & Indicator, denoting whether the driver has been matched to a request before the next drop-off \\ \hline
		$s$ & State, $s=(l,t,I)$ \\ \hline
		$S$ & State space, a collection of all states \\ \hline
		$a$ & Action \\ \hline
		$A$ & Action space \\ \hline
		$t_{seek}(l_a)$ & Time spent on seeking for a passenger in grid $l_a$ \\ \hline
		$t_{drive}(l,k)$ & Time spent on moving from grid $l$ to grid $k$ \\ \hline
		$d_{seek}(l_a)$ & Distance traveled when seeking for a passenger in grid $l$ \\ \hline
		$d_{drive}(l,k)$ & Distance traveled for moving from grid $l$ to grid $k$ \\ \hline
		$p_{order\_match}(l_a)$ & The probability that the driver can be matched to a request during cruising in grid $l_a$ \\ \hline
		$p_{pickup}(l_a,l'')$ & The probability of picking up a passenger in grid $l''$ when the request from the passenger was matched to the driver in grid $l_a$\\ \hline	
		$p_{dest}(l'',l''')$ & The probability of dropping off a passenger in grid $l'''$ when the passenger was picked up in grid $l''$\\ \hline	
		$p_{match}(l'',l''')$ & The probability of receiving a new request before the driver finishing her current order at grid $l'''$\\ \hline
		$f(l'',l''')$ & The average taxi fare from grid $l''$ to grid $l'''$\\ \hline
		$\alpha$ & Coefficient of fuel consumption and other operating costs per unit distance\\ \hline
	\end{tabular}
\end{table}

\subsubsection{States}

In our MDP model, the state $s=(l, t, I)$ consists of three components, namely, a grid index $l \in L$, current time $t \in T$, and an indicator $I \in \{0, 1\}$. Note that a hexagonal grid world setting with 6,421 grids is adopted in this research and will be explained later. Considering the fact that an e-hailing driver may receive a request before she drops off the previous passenger, we have therefore added an indicator into the state. The indicator denotes whether the driver has been matched to a request before she arrives at the current state. Accordingly, states with indicator $0$ are decision-making states in which the driver needs to spend time on seeking the next passenger, and states with indicator $1$ are non-decision-making states. For example, $(1, 2, 1)$ is a non-decision-making state which says that the driver is in grid $1$ when $t=2$ and the driver has already been matched to a request so she will not spend time on seeking at the current state. 

\subsubsection{Actions}

In decision-making states, the driver has to choose one from eight allowable actions, denoted as $A$. In non-decision-making states, the driver will not take any action but drive to pick up the next passenger and transport the passenger to the destination. Among the allowable action space $A$, each of the first six actions is to transit from the current grid to one of the six neighbor grids. Note that some of the six neighboring grids may be non-reachable, we thus add a large penalty, i.e., a large distance, to the transition from a grid to a non-reachable neighboring grid to prevent the agent from taking the action which leads the agent to the non-reachable neighboring grid. The seventh action is to stay and cruise around within the current grid. The last action is to wait in the current grid. We stress that the last two actions are essentially different because from the data we have observed that some drivers will just wait near the previous drop-off spot, especially when they are around downtown or transportation terminals while some drivers usually cruise within the current grid after completing a ride. In addition, the fuel cost associated with waiting can be neglected while that of staying can be substantial because the driver keeps cruising around during his/her staying in the current grid. %Specifically, we take the distance traveled during staying as 0.5 kilometers considering the size of the hexagonal grid (approximately 700 meters) in our model. 
Furthermore, drivers can take a rest and refresh their minds during waiting and hence their driving strategy can be more efficient for future trips. These arguments, however, do not necessarily suggest that the driver should always choose waiting rather than staying. Actually, drivers have to cruise around to get closer to the potential requests under certain circumstances. 

\subsubsection{State transition}

After completing a ride, there are two possible scenarios according to two different values of the indicator. If the indicator is $0$, the driver needs to specify an action, i.e. where to find the next passenger, and then moves into the grid along the direction defined by the action and spends some amount of time seeking for the next passenger in the new grid. There are two possible outcomes associated with this passenger seeking process. Either the driver confirms a request and ends up arriving at a different grid by following the passenger's travel plan or the driver fails to find a request and stays in the current grid. The reward is usually positive for the former while negative for the latter due to the fuel consumption and other operating costs. If the indicator is $1$, the driver drives to the pick-up spot and then transports the passenger to the destination without any passenger seeking involved.

\begin{figure}[H]
	\centering
	\includegraphics[width=0.99\linewidth,height=\textheight,keepaspectratio]{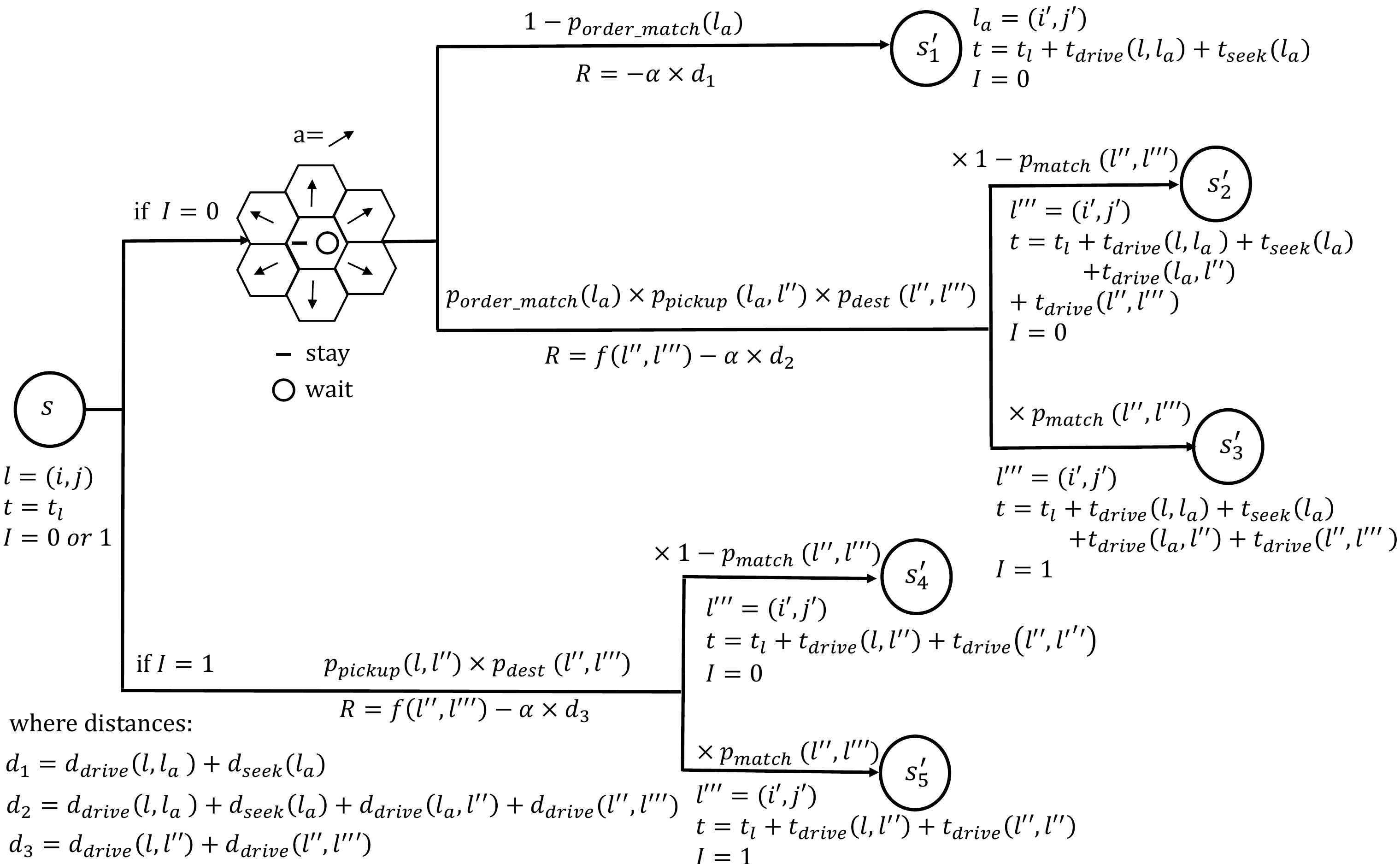}
	\centering 
	\caption{MDP state transition}
	\label{fig:state_transition}
\end{figure}

Figure~(\ref{fig:state_transition}) illustrates the aforementioned state transition process. The driver currently stays at state $s = (l, t, I)$. 

If $I=0$, the driver specifies an action $a$, which is assumably taken as going northeast in the demonstration. Then the driver moves into the grid $l_a$ along the direction defined by the action $a$ and thereafter spends some amount of time $t_{seek}(l_a)$ on seeking the next passenger. There are two possible outcomes of this passenger seeking process. 

The first possibility is that the driver fails to get any request in grid $l_a$ after $t_{seek}(l_a)$. In this case, the state of the driver will be $s'_1=(l', t+t_{drive}(l,l_a)+t_{seek}(l_a), 0)$. The reward for this passenger seeking process is $R=-\alpha(d_{drive}(l,l_a) + d_{seek}(l_a))$, which is negative. Let $P_{order\_match}(l_a)$ denote the probability that the driver will receive at least one request in grid $l_a$. Then the probability of the occurrence of this outcome is $1-P_{order\_match}(l_a)$. In other words, with probability $1-P_{order\_match}(l_a)$, the driver will end up in state $s'_1=(l_a, t+t_{drive}(l,l_a)+t_{seek}(l_a), 0)$.

The second possibility is that the driver confirms one request during the cruising process in $l_a$. The probability of the occurrence of this outcome is $P_{order\_match}(l_a)$. We let $P_{pickup}(l_a,l'')$ denote the probability of confirming a request in grid $l_a$ and picking up the passenger in grid $l''$. Once the passenger is on board, the driver will directly move to the destination $l'''$, which only depends on the passenger's travel plan. We let $P_{dest}(l'',l''')$ denote the probability of picking up a passenger in grid $l''$ and dropping off the passenger in grid $l'''$. After dropping off the passenger, the driver will end up in grid $l'''$ at time $t+t_{drive}(l,l_a)+t_{seek}(l_a)+t_{drive}(l_a,l'')+t_{drive}(l'',l''')$ and earn a reward of $r(l,l''')=f(l'',l''') - \alpha (d_{drive}(l,l_a) + d_{seek}(l_a) + d_{drive}(l_a,l'') + d_{drive}(l'',l'''))$. Hence, the probability of the driver receives a request in grid $l_a$, picks up the passenger in grid $l''$, and transports the passenger to grid $l'''$ is $P_{order\_match}(l_a) \times P_{pickup}(l_a,l'') \times P_{dest}(l'',l''')$. Notice that for a driver, during her trip from the passenger's origin $l''$ to the passenger's destination $l'''$, there is a probability at which the driver will confirm a request before she drops off the passenger. Let $P_{match}(l'',l''')$ denote the probability of receiving a request before the driver reaches grid $l'''$. Then we can conclude that with probability $P_{order\_match}(l_a)\times P_{pickup}(l_a,l'') \times P_{dest}(l'',l''') \times (1-P_{match}(l'',l'''))$, the driver will end up in state $s'_2=(l''', t+t_{drive}(l,l_a)+t_{seek}(l_a)+t_{drive}(l_a,l'')+t_{drive}(l'',l'''), 0)$, and with probability $P_{order\_match}(l_a)\times P_{pickup}(l_a,l'') \times P_{dest}(l'',l''') \times P_{match}(l'',l''')$, the driver will end up in state $s'_3=(l''', t+t_{drive}(l,l_a)+t_{seek}(l_a)+t_{drive}(l_a,l'')+t_{drive}(l'',l'''), 1)$.

If $I=1$, the driver will not need to specify any action and will directly drive to the pick-up spot of the next passenger and then transport the passenger to the destination. Again, during her trip to the passenger's destination, there is a probability, denoted as $P_{match}(l'',l''')$, at which the driver will receive a request before she drops off the passenger. As illustrated in Figure~(\ref{fig:state_transition}), with probability $P_{pickup}(l,l'') \times P_{dest}(l'',l''') \times (1-P_{match}(l'',l'''))$, the driver will end up in state $s'_4=(l''', t+t_{drive}(l,l'')+t_{drive}(l'',l'''), 0)$; with probability $P_{pickup}(l,l'') \times P_{dest}(l'',l''') \times P_{match}(l'',l''')$, the driver will end up in state $s'_5=(l''', t+t_{drive}(l,l'')+t_{drive}(l'',l'''), 1)$.

In both scenarios, namely, either $I=0$ or $I=1$, the driver will thereafter start the whole process from $s'$ again until the current time exceeds the time interval, i.e., a terminal state has been reached.

\subsubsection{Solving MDP} \label{sec:objective}

The objective of the MDP model is to maximize the total expected revenue of a driver. Considering the fact that in a time interval, a driver can finish a finite number of pick-up and drop-off cycles, indicating that the MDP model is finite-horizon. When current time of the driver has reached the end of the time interval, no more actions can be taken and no more rewards can be earned. Suppose a driver is currently at state $s=(l,t,I)$. If $I=0$, meaning that the driver is at a decision-making state, the maximum expected revenue that a driver can earn by starting from $s$ and specifying an action $a$ is

\begin{eqnarray}
Q(s, a) & = & (1- P_{order\_match}(l_a)) \times [-\alpha(d_{drive}(l,l_a) + d_{seek}(l_a)) + V^*(s'_1)] \nonumber\\ 
& \qquad & + \sum_{l'' \in L} \sum_{l''' \in L}P_{order\_match}(l_a)\times P_{pickup}(l_a,l'') \times P_{dest}(l'',l''') \nonumber \\
& \qquad & \qquad \times [f(l'',l''') - \alpha ( d_{drive}(l,l_a) + d_{seek}(l_a) + d_{drive}(l_a,l'') + d_{drive}(l'',l''')) \nonumber \\
& \qquad & \qquad \quad + (1-P_{match}(l''')) \times V^*(s'_2) + P_{match}(l''') \times V^*(s'_3)]  \label{eqn:maximum_reward}
\end{eqnarray}
where $l_a = l_a(s,a)$, meaning that the grid $l_a$ in which an e-hailing driver will be cruising is dependent on the current state $s$, actually through the grid index $l$ of $s$, and the specified action $a$, $s'_1=(l_a, t+t_{drive}(l,l_a)+t_{seek}(l_a), 0)$, $s'_2=(l''', t+t_{drive}(l,l_a)+t_{seek}(l_a)+t_{drive}(l_a,l'')+t_{drive}(l'',l'''), 0)$, $s'_3=(l''', t+t_{drive}(l,l_a)+t_{seek}(l_a)+t_{drive}(l_a,l'')+t_{drive}(l'',l'''), 1)$ , and $V^*(s'_1)$, $V^*(s'_2)$, and $V^*(s'_3)$ stand for the maximum expected revenue that a driver can earn by reaching state $s'_1$, $s'_2$, and $s'_3$, respectively. 
If $I=1$, meaning that the driver is at a non-decision-making state, the driver will not specify any action, and the expected revenue that the driver can earn is

\begin{eqnarray}
Q(s, .) & = & \sum_{l'' \in L} \sum_{l''' \in L} P_{pickup}(l,l'') \times P_{dest}(l'',l''') \nonumber \\ 
& \qquad &\times [f(l'',l''') - \alpha (d_{drive}(l,l'') + d_{drive}(l'',l''')) \nonumber \\
& \qquad & + (1-P_{match}(l'',l''')) \times V^*(s'_4) + P_{match}(l'',l''') \times V^*(s'_5)] \label{eqn:maximum_reward_p2}
\end{eqnarray}
where $s'_4=(l''', t+t_{drive}(l,l'')+t_{drive}(l'',l'''), 0)$, $s'_5=(l''', t+t_{drive}(l,l'')+t_{drive}(l'',l'''), 1)$ , and $V^*(s'_4)$ and $V^*(s'_5)$ stand for the maximum expected revenue that a driver can earn by reaching state $s'_4$ and $s'_5$, respectively.

Then the optimal policy for a driver to follow at a decision-making state $s$ is 
\begin{equation}
\pi(s) = argmax_a[Q(s, a)],
\label{eqn:policy}
\end{equation} 
and the maximum expected revenue that a driver can earn by reaching state $s$ is
\begin{equation}
V^*(s) = \left\{
\begin{array}{ll}
max_aQ(s, a) & \text{if $s$ is a decision-making state} \\
Q(s,.) & \text{if $s$ is a non-decision-making state}\\
\end{array} 
\right. 
\label{eqn:optimal_v}
\end{equation}

The policy in Equation~(\ref{eqn:policy}) is deterministic, meaning that the driver can only take one action at the current decision-making state $s$ if she follows the policy. Actually here we slightly abuse the notation. The policy is supposed to be $\pi(argmax_a[Q(s, a)]|s) = 1$, i.e., the probability of taking action $argmax_a[Q(s, a)]$ at state $s$ is 1. It is equivalent to say that at state $s$, the action to take is $argmax_a[Q(s, a)]$, and thus we write the policy at state $s$ as Equation~(\ref{eqn:policy}). A deterministic policy defines a one-to-one mapping from a state to an action. The deterministic policy works when there is only one driver who learns the optimal policy and follows the policy. Otherwise, there might be excess taxi supply at some areas, resulting in a localized competition among taxis. A circulating mechanism was employed to tackle this overload problem \citep{ge_energy-efficient_2010}. A multi-agent reinforcement learning approach \citep{lin_efficient_2018} was proposed to consider the competition among drivers. In this research, we use a dynamic adjustment strategy to update the order matching probability when multiple idling e-hailing drivers are guided into the same grid. The proposed dynamic adjustment strategy will be introduced in Section~(\ref{sec:multi_mdp}). %In this research, we use a randomized optimal policy instead of the deterministic policy to alleviate the localized competition among taxis. A randomized policy is a mapping from a decision-making state to a distribution over actions.
%\begin{equation}
%prob(\pi(s) = a) = \dfrac{e^{\beta \times Q(s, a)}}{\sum_{a' \in A} e^{\beta \times Q(s, a')} }, ~ \forall a \in A
%\label{eqn:s_policy}
%\end{equation}
%where $\beta$ is a temperature parameter, which determines to what extent we want the randomized policy to align with the deterministic policy.  The larger the $\beta$, the closer the randomized policy to the deterministic policy.
%The corresponding maximum expected revenue that a driver can earn by reaching state $s$ is
%\begin{equation}
%V^*(s) = \left\{
%\begin{array}{ll}
%\sum_{a \in A} prob(\pi(s) = a) Q(s, a) & \text{if $s$ is a decision-making state} \\
%Q(s,.) & \text{if $s$ is a non-decision-making state}\\
%\end{array} 
%\right. 
%\label{eqn:s_optimal_v}
%\end{equation}

%Obviously, the randomized policy shown in Equation~(\ref{eqn:s_optimal_v}) will guide drivers to different directions even though they follow the exact same policy. This will alleviate the aforementioned localized competition among drivers to some degree since less drivers will be referred to the same location compared with the case in which deterministic policy is used. In other words, the competition among several drivers will still exist at some locations, but the degree of the competition will be lower. Thereafter we further assume the probability of getting an order match in a grid is independent of the number of taxi being present in the grid.

To efficiently solve the MDP, i.e., to derive an optimal policy, a dynamic programming approach is employed \citep{Bertsekas_2000, sutton_introduction_1998}. The basic idea of the dynamic programming algorithm is to divide the overall problem into subproblems and hence to make use of the results of the subproblems to solve the overall problem. An important advantage of the dynamic programming algorithm is that it caches results of all subproblems and thus it is guaranteed that the same subproblem is only solved once.

Now we elucidate how we apply the dynamic programming algorithm to solve the MDP. The goal is to solve the optimal value for all states $s = (l,t,I)$, where $l \in L$, $t \in \{0,1,2,\cdot, 180\}$, and $I \in \{0, 1\}$. There are in total $6,421 \times 181 \times 2 = 2,324,402$ states, and half of them are decision-making states. We define one subproblem as solving the optimal value for one state and thus we have in total $2,324,402$ subproblems. Noticing that at the final time step, i.e., $t=T=180$, the maximum expected reward that a driver can earn is obviously zero, we thus have $V^*(s) = 0$ for all states $s$ where $t = T$. For any state $s$ with $t<T$ and a chosen action $a$, the calculation of the state-action value $Q(s,a)$ depends on the value of some future states, i.e., $s'_1$, $s'_2$, and $s'_3$ in Equation~(\ref{eqn:maximum_reward}). In other words, the subproblem, i.e., solving the optimal value for state $s$, depends on some subproblems, i.e., solving the optimal value of some future states, e.g., $s'_1$, $s'_2$, and $s'_3$. For a future state $s'$, there might be several states $s$ from which the agent will reach the future state $s'$, indicating the calculation of the optimal value of all these states $s$ requires the calculation of the optimal value of the future state $s'$, resulting in calculating the optimal value of the same state $s'$ multiple times and thus wasting computation power. To avoid the repeated calculation of the optimal value for the same state, we adopt the dynamic programming algorithm. Since the optimal values for all states with $t=T$ are known and the optimal value of a state $s$ depends on the optimal value of some future states, we solve the optimal value of states backwards in time and simply store the solved optimal values in a hash table. Then for a state $s$ and a chosen action $a$, we simply read the optimal values of future states $s'_1$, $s'_2$, and $s'_3$ from the hash table and use Equation~(\ref{eqn:maximum_reward}) to calculate the state-action value $Q(s,a)$, based on which the optimal value of the state $s$ can be derived from Equation~(\ref{eqn:optimal_v}). The pesudo code is in Algorithm~(\ref{alg:dynamic_prog}).

\begin{algorithm}[H]
	\caption{Dynamic programming algorithm}
	\label{alg:dynamic_prog}
	\begin{algorithmic}[1]
		\State{Input: $L$, $T=180$, $A$, $P_{order\_match}$, $P_{pickup}$, $P_{dest}$, $P_{match}$, fare $f$, $d_{drive}$, $d_{seek}$, $t_{drive}$, $t_{seek}$}
		\State \textbf{Initialize:} a hash table $V^*$ to store optimal values
		\State $V^*(s)=0$ for all states $s$ where $t=T$
		\For{\texttt{$t=T-1$ to 1}}
		\For{\texttt{$l \in L$}}
		\State Form a decision-making state $s=(l,t,I=0)$
		\For{\texttt{$a \in A$}}
		\State Calculate $Q(s,a)$ by Equation~(\ref{eqn:maximum_reward}), the optimal values of the dependent future states are read from the hash table $V^*$
		\EndFor
		\State Derive the optimal policy for decision-making state $s$, $\pi(s)$, by Equation~(\ref{eqn:policy})
		\State Calculate the optimal value for state $s$, $V^*(s)$, by Equation~(\ref{eqn:optimal_v})
		\State  
		\State Form a non-decision-making state $s=(l,t,I=1)$
		\State Calculate $Q(s,.)$ by Equation~(\ref{eqn:maximum_reward_p2}), the optimal values of the dependent future states are read from the hash table $V^*$
		\State Calculate the optimal value for state $s$, $V^*(s)$, by Equation~(\ref{eqn:optimal_v})
		\EndFor
		\EndFor
		\State return the $V^*$ and $\pi$
	\end{algorithmic}
\end{algorithm}

\subsection{Extracting parameters from data} \label{subsec:parameters}
In the dataset we used in this research, we have GPS trajectories for both the empty and occupied trips. We now introduce how to extract the parameters we used in the state transition from the dataset. 

\subsubsection{Order matching probability $P_{order\_match}$}
The order matching probability estimates the probability at which a vacant taxi can be matched to a passenger when the taxi is cruising, including staying, or waiting at grid $l_a$. As we have mentioned before, the purposes for introducing waiting and staying in this work are different. In addition to six actions which allow an e-hailing driver to move into one of the six neighboring grids, the action staying gives the driver extra flexibility in choosing to stay and cruise within the current grid due to some potential benefits, such as a relatively high order matching probability in the current grid, a possibly high cost to move into neighboring grids vacantly, etc. Actually, as we have listed in Table~(\ref{tab:mdp}), there are several studies in the literature that have already included the action staying into the action space, such as \citep{rong_rich_2016}, \citep{verma_augmenting_2017}, and \citep{lin_efficient_2018}. Thus, the way to calculate the order matching probability for staying is the same as the way to calculate the order matching probability for cruising into one of the six neighboring grids. In other words, the order matching probability for a driver just entering the grid from one of the six neighboring grids is supposed to be the same as the order matching probability for a driver who was in the grid and chose to stay in the grid.

Waiting, different from staying and other six actions which allow the driver to move into one of the six neighboring grids, is included into the action space based on the observation that sometimes a driver will choose to stop cruising and simply to wait statically for passenger requests to come in, especially when the driver is around downtown or transportation terminals. The action waiting was previously included in the action space in \citep{gao_optimize_2018}.

We thus approximate the order matching probabilities for cruising and waiting separately. We say a driver is waiting for a passenger request whenever the driver's traveling distance is less than 200 meters for a 3-minute interval. To rule out some unrealistic waiting actions, such as a driver being stuck in traffic, we further limit the possible locations for waiting to be the places around subway stations, bus terminals, airports, and some famous tourism attractions. For cruising, the order matching probability can be approximated as the ratio of the number of times that a taxi is matched to a passenger in grid $l_a$ while cruising, denoted as $n_{order\_match\_cruising}(l_a)$, to the number of times that the grid $l_a$ is passed by an empty taxi while cruising, denoted as $n_{passby\_cruising}(l_a)$. For waiting, the order matching probability can be approximated as the ratio of the number of times that a taxi is matched to a passenger in grid $l_a$ while waiting, denoted as $n_{order\_match\_waiting}(l_a)$, to the number of times that empty taxis have waited in the grid $l_a$, denoted as $n_{passby\_waiting}(l_a)$.

\begin{equation}
	P_{order\_match}(l_a) = \left\{
	\begin{array}{ll}
	\dfrac{n_{order\_match\_cruising}(l_a)}{n_{passby\_cruising}(l_a)} & cruising \\
	\dfrac{n_{order\_match\_waiting}(l_a)}{n_{passby\_waiting}(l_a)} & waiting\\
	\end{array} 
	\right. 
	\label{eqn:order_match}
\end{equation}

\subsubsection{Pick-up probability $P_{pickup}$}
The pick-up probability $P_{pickup}(l_a,l'')$ measures the likelihood of picking up a passenger at grid $l''$ when the request sent from the passenger was matched to the driver at grid $l_a$. This parameter can be estimated as the ratio of the the number of passenger pick-ups in grid $l''$ which were matched to drivers in grid $l_a$, denoted as $n_{pickup}(l_a,l'')$, to $n_{order\_match}(l_a)$, which is the summation of $n_{order\_match\_cruising}(l_a)$ and $n_{order\_match\_waiting}(l_a)$.
\begin{equation}
	P_{pickup}(l_a, l'') = \dfrac{n_{pickup}(l_a,l'')}{n_{order\_match}(l_a)}
	\label{eqn:pickup}
\end{equation}

\subsubsection{Destination probability $P_{dest}$}
The destination probability $P_{dest}(l'',l''')$ measures the likelihood of the destination of the passenger being grid $l'''$ when the passenger was picked up in grid $l''$. This parameter can be estimated by dividing the number of trips ending in grid $l'''$ which originated from grid $l''$, denote as $n_{dest}(l'',l''')$, by the total number of pick-ups in grid $l''$, denoted as $n_{pickup}(l'')$.
\begin{equation}
	P_{dest}(l'', l''') = \dfrac{n_{dest}(l'',l''')}{n_{pickup}(l'')}
	\label{eqn:dest}
\end{equation}

\subsubsection{Order matching probability while on trip $P_{match}$} 

As we have mentioned before, there is a probability at which the driver will receive a request when she is on the trip to transport the current passenger to the destination. We denote this order matching probability while on trip as $P_{match}$. This probability can be estimated by dividing the number of occupied trips among which there is at least one request received by the driver before the driver reaching the destination $l'''$ while the origin is $l''$, denoted as $n_{match}(l'',l''')$, by the total number of occupied trips ending in grid $l'''$ and originating in grid $l''$, denoted as $n_{trips}(l'',l''')$.
\begin{equation}
	P_{match}(l'',l''') = \dfrac{n_{match}(l'',l''')}{n_{trips}(l'',l''')}
	\label{eqn:match}
\end{equation}

\subsubsection{Driving time $t_{drive}$ and driving distance $d_{drive}$}

The driving time $t_{drive}(l, k)$ and the driving distance $d_{drive}(l,k)$ denote the estimated driving time and driving distance from grid $l$ to grid $k$, respectively. Here we simply take the average of all driving times from grid $l$ to grid $k$ as an approximation of the $t_{drive}(l,k)$. Similarly, the driving distance is calculated by taking the average of all driving distances between grid $l$ and grid $k$. 

\subsubsection{Taxi fare $f$}

The taxi fare $f(l'', l''')$ denotes the estimated gross revenue that a driver can earn by transporting a passenger from grid $l''$ to her destination grid $l'''$. Here we take the average of all the fares of the occupied trips which are from grid $l''$ to grid $l'''$ as a proxy of the real taxi fare from grid $l''$ to $l'''$.

\subsubsection{Seeking time $t_{seek}$ and seeking distance $d_{seek}$} 

\begin{figure}[H]
	\centering
	\includegraphics[width=0.7\linewidth,height=\textheight,keepaspectratio]{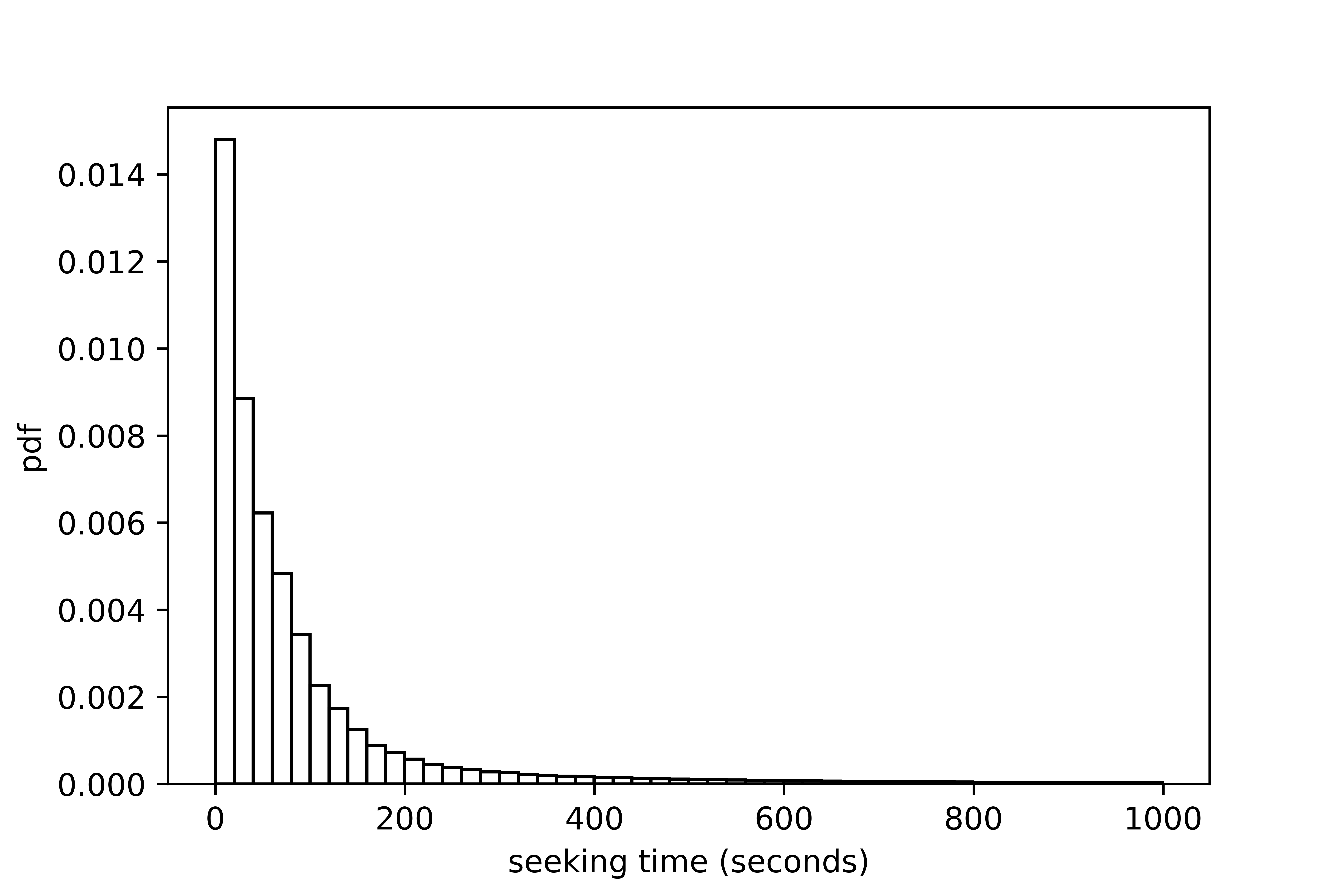}
	\centering 
	\caption{Distribution of seeking time}
	\label{fig:seeking}
\end{figure}

The seeking time $t_{seek}(l_a)$ and the seeking distance $d_{seek}(l_a)$ denote the estimated seeking time and seeking distance within grid $l_a$, respectively. From the field data, the distribution of the seeking time in each grid was extracted and is shown in Figure~(\ref{fig:seeking}). The median of the distribution of the seeking time is approximately 45 seconds. Since the time step size is 1 minute in this work, thus we simply take the seeking time as 1 minute. Considering the average speed of seeking trips (around 300 meters/minute), the seeking distance is taken as 300 meters for each grid. Note that the seeking distance is zero when the driver chooses to wait.

\subsubsection{The fuel consumption coefficient $\alpha$}

$\alpha$ estimates the fuel consumption and other operating cost per unit distance during driving. In the literature, the value of $\alpha$ is typically assumed to be a known constant based on some common knowledge \citep{rong_rich_2016,verma_augmenting_2017, lin_efficient_2018, yu_markov_2019}. This assumption, however, can easily result in a gap between the reward function in MDP models and the reward function of real drivers, meaning that drivers will not follow the optimal policy since the reward function used to derive the optimal policy is not the reward function of real drivers. To more appropriately determine $\alpha$, we opt for the inverse reinforcement learning (IRL) approach. IRL is a powerful technique to disclose the underlying reward function based on the observed behaviors, especially in the field of imitation learning where an agent's behavior is observed by a learner who tries to imitate the agent \citep{ziebart_maximum_2008}.  

A linear programming formulation of IRL in finite state spaces was first proposed in \citep{ng_algorithms_2000}, where an extension to large state spaces was also made possible by adopting a linear function approximation. In the context of understanding the observed behaviors, \cite{liu_understanding_2013} argued that a mixed integer linear programming formulation can be more revealing. Noticing the similarity between the two approaches, we take the mixed integer linear programming formulation for simplicity. To make the paper self-explanatory, we provide a brief introduction of mixed integer linear programming formulation. Interested readers are referred to \citep{liu_understanding_2013} for detailed explanations.

In the linear programming formulation, the underlying reward function $R(s, s')$ from a state $s$ to another state $s'$ is expressed as a linear combination of some simple known reward functions $\phi_i$s, i.e.,
\begin{equation}
R(s, s') = \alpha_1 \phi_1(s,s') + \alpha_2 \phi_2(s,s') + \cdots + \alpha_n \phi_n(s,s')
\end{equation}
As an example, $\phi_i(s, s')$ can be either the fare a driver can collect from $s$ to $s'$ or the distance a driver traveled from $s$ to $s'$. The former is considered to be positive while the latter is negative. For each simple reward function $\phi_i$, the optimal value function $V^{\phi_i}$ can be derived by solving the MDP with $\phi_i$ as the reward function. Due to linearity, the optimal value function under the underlying reward function can be calculated as
\begin{equation}
V^{R} = \alpha_1 V^{\phi_1} + \alpha_2 V^{\phi_2} + \cdots + \alpha_n V^{\phi_n}
\end{equation}
The optimal policy $\pi^R$ is also derived when solving the MDP. The objective of the mixed integer linear programming IRL is to minimize the difference between the optimal policy $\pi^R$ and the observed policy $\pi^O$, i.e., minimize$\sum_{s \in S} [\pi^R(s) \neq \pi^O(s)]$. After some mathematical manipulations, a mixed integer linear programming formulation is formed as
\begin{eqnarray}
& \text{minimize} &  \sum_{s \in S} C_s \\
& \text{s.t. } & \nonumber\\
& & \alpha_i \geq 0 \label{eqn:c1} \\
%& & \sum_i \alpha_i = 1 \label{eqn:c2}\\
& & \sum_{s'}P_{\pi^O}(s,s') (R(s,s') + \gamma V^{\pi^R}(s')) \nonumber\\ 
& & \qquad - \sum_{s'}P_{a}(s,s') (R(s,s') + \gamma V^{\pi^R}(s'))+ M\times C_s \geq 0 \label{eqn:c3}
\end{eqnarray}
where $C_s$ is a binary variable and takes value of 0 or 1 and $M$ is an arbitrarily large number. $\sum_{s \in S} C_s$ denotes the number of states where the observed policy and the optimal policy differ. The objective is thus to minimize the difference between $\pi^R$ and $\pi^O$. Constraint~(\ref{eqn:c1}) restricts the weighting parameter to be nonnegative. % within the range of $[0, 1]$.%, and constraint~(\ref{eqn:c2}) requires the summation of all weighting parameters to be 1. 
In the last constraint~(\ref{eqn:c3}), the first term $\sum_{s'}P_{\pi^O}(s,s') (R(s,s') + \gamma V^{\pi^R}(s'))$ is the optimal value at state $s$ following the observed policy, and the second term $\sum_{s'}P_{a}(s,s') (R(s,s') + \gamma V^{\pi^R}(s'))$ is the value following other policies. Thus, we expect $\sum_{s'}P_{\pi^O}(s,s') (R(s,s') + \gamma V^{\pi^R}(s')) \geq\sum_{s'}P_{a}(s,s') (R(s,s') + \gamma V^{\pi^R}(s'))$ to hold. In reality, however, this constraint can be violated, and thus we add a large positive number to the violated constraint to keep (\ref{eqn:c3}) hold. $P_{\pi}(s,s')$ is the probability of the transition $s \rightarrow s'$ following the policy $\pi$. In this work, $P_{\pi}(s,s')$ or $P_{a}(s,s')$ has been demonstrated in Figure~(\ref{fig:state_transition}). 

\begin{rem}
	To identify an overall reward function (i.e., determining $\alpha_i$s), we assume part of the observed policy is optimal or near optimal. %This is the major assumption in this study. 
	Different from simply assuming a reward function based on common knowledge in the literature \citep{rong_rich_2016,verma_augmenting_2017, lin_efficient_2018, yu_markov_2019}, we argue that the assumption used here is relaxed to some degree. Real drivers, at least part of them, are deemed to be intelligent and experienced and exhibit optimal or near optimal strategies. For a state visited by multiple drivers for multiple times, we believe the most frequently taken action carries useful information about the optimal strategy in this state and reflects the crowd wisdom. After uncovering the underlying reward function, the purpose of solving the MDP and thus deriving the optimal policy is to complete and correct the optimal policy. The incompleteness and noise in the observed policy stem from the following aspects: (1) For a real-world problem with a huge number of states (in our case study in Section 4 the number of decision-making states is $6,421 \times 180 = 1,115,780$), a considerable portion of the states are not visited or at least not frequently visited; (2) Even in states with sufficient data (i.e., enough actions chosen by agents in this state), the most frequently chosen action which is taken as the observed policy in this state can still differ from the subsequent derived optimal policy, due to behavioral inconsistency; and (3) the observed policy, which is assumed to be deterministic, can be ambiguous when several actions share similar frequency in some states.
\end{rem} 

%The observed policy $\pi^O(s)$ in each state $s$ is simply the most frequently taken action by all drivers in that state. Applying the aforementioned IRL technique with two known reward functions (i.e., fare $\phi_1(s,s')$ and traveling distance $\phi_2(s,s')$), we obtain two coefficients $\alpha_1 = 0.61$ and $\alpha_2 = 0.39$.

%Here we take $\alpha=0.5$ (Yuan/kilometer). 

\subsection{Numerical example}

\begin{figure}[H]
	\centering 
	\subfloat[Setup]{\includegraphics[scale=.43]{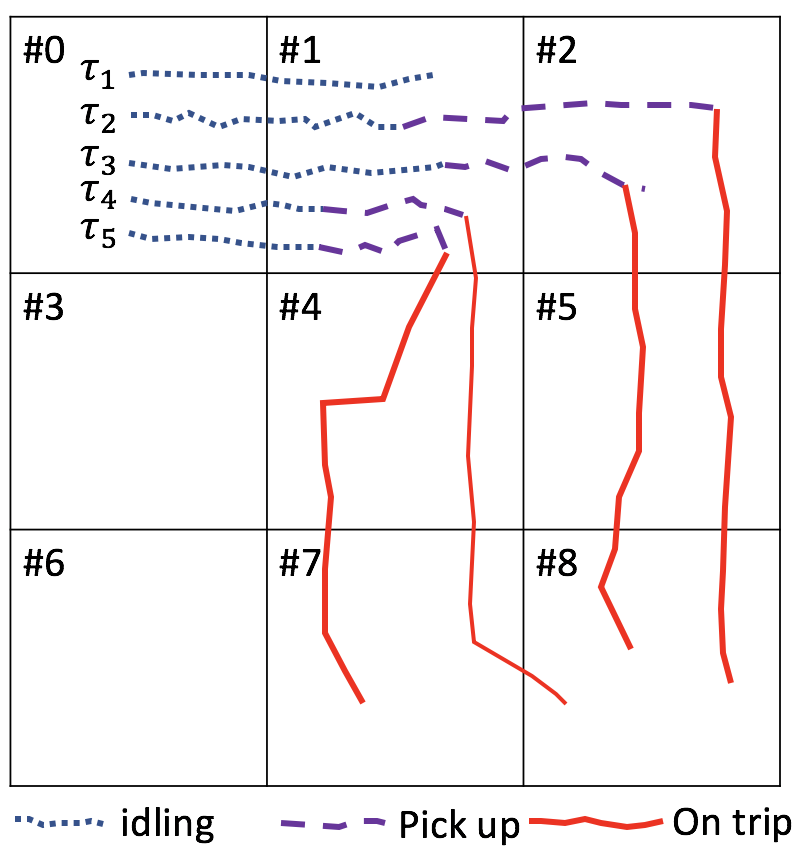}\label{subfig:setup}}~
	\subfloat[State transition]{\includegraphics[scale=.57]{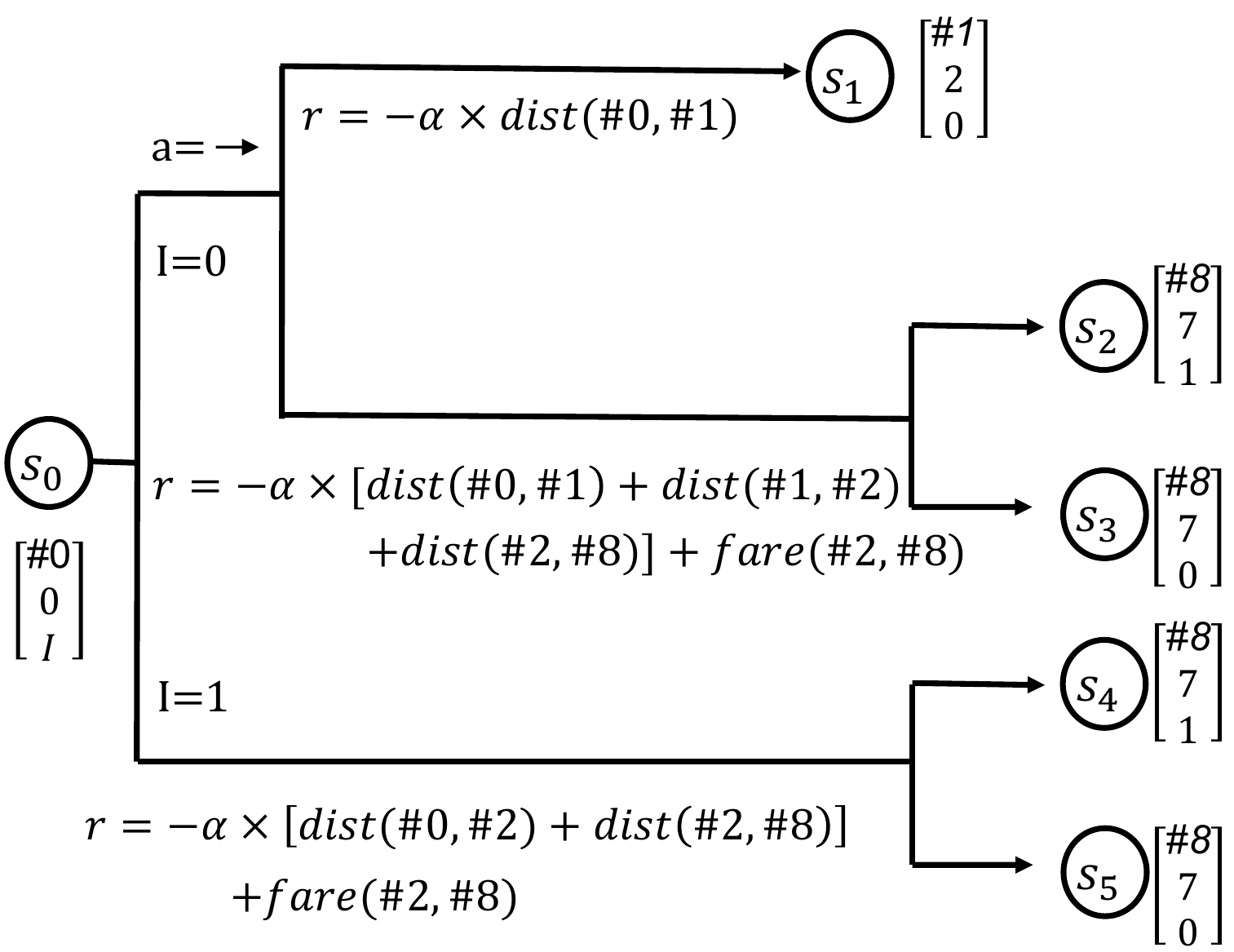}\label{subfig:state_transition_small}} 
	\caption{Setup of the small grid world example and the corresponding state transition}
	\label{fig:comparison} 
\end{figure}

To illustrate the Markov Decision Process of e-hailing drivers, we use a 3 by 3 grid world numerical example, as shown in Figure~(\ref{subfig:setup}). 

Suppose now we have the following five trajectories.
\begin{enumerate}
	\item $\tau_1 = $ $(\#0, 0, 0)$ $\xrightarrow{\text{idling}}$ $(\#1, 2, 0)$
	\item $\tau_2 = $ $(\#0, 0, 0)$ $\xrightarrow{\text{idling}}$ $(\#1, 2, 0)$ $\xrightarrow{\text{pickup}}$ $(\#2, 3, 0)$ $\xrightarrow[\text{no match}]{\text{ontrip}}$ $(\#8, 7, 0)$
	\item $\tau_3 = $ $(\#0, 0, 0)$ $\xrightarrow{\text{idling}}$ $(\#1, 2, 0)$ $\xrightarrow{\text{pickup}}$ $(\#2, 3, 0)$ $\xrightarrow[\text{match}]{\text{ontrip}}$ $(\#8, 7, 1)$
	\item $\tau_4 = $ $(\#0, 0, 0)$ $\xrightarrow{\text{idling}}$ $(\#1, 2, 0)$ $\xrightarrow{\text{pickup}}$ $(\#1, 3, 0)$ $\xrightarrow[\text{no match}]{\text{ontrip}}$ $(\#8, 7, 0)$
	\item $\tau_5 = $ $(\#0, 0, 0)$ $\xrightarrow{\text{idling}}$ $(\#1, 2, 0)$ $\xrightarrow{\text{pickup}}$ $(\#1, 3, 0)$ $\xrightarrow[\text{no match}]{\text{ontrip}}$ $(\#7, 6, 0)$
\end{enumerate}
Each element in the trajectory is a tuple consisting of three items, namely, the grid index, current time, and a status indicator showing if the driver has been matched to another order during the trip. For example, $(\#0, 0, 0)$ basically states that the driver is at grid $\#0$ at time $0$, and the driver has not been matched to any order before she finished the previous trip. 

All five trajectories started in grid $\#0$, and then the driver moved into grid $\#1$ during idling. After the driver searched the grid $\#1$, there are two possible outcomes: either the driver finds an order match or the driver fails to find any e-hailing order. If the driver fails to get an order match after searching, the driver will move into other grids to find another order or the driver will stop working. To simplify the demonstration, we simply assume the trajectory $\tau_1$ ends in grid $\#1$. For other four trajectories, the driver managed to find an e-hailing order in grid $\#1$. Based on this piece information, we can calculate the probability of finding an e-hailing order in grid $\#1$ as $P_{order\_match}(\#1) = \dfrac{n_{order\_match}(\#1)}{n_{passby}(\#1)} = \dfrac{4}{5} = 80\%$. 

After confirming an order match in grid $\#1$, the driver drives into grid $\#2$ to pick up the passenger in trajectories $\tau_2$ and $\tau_3$ and stays within grid $\#1$ to pick up the passenger in trajectories $\tau_4$ and $\tau_5$ , respectively. Thus, the pick-up probability can be calculated as $P_{pickup}(\#1, \#1) = \dfrac{n_{pickup}(\#1, \#1)}{n_{order\_match}(\#1)} = \dfrac{2}{4} = 50\%$ and $P_{pickup}(\#1, \#2) = \dfrac{n_{pickup}(\#1, \#2)}{n_{order\_match}(\#1)} = \dfrac{2}{4} = 50\%$. 

When the driver picks up the passenger in grid $\#1$, as illustrated in trajectories $\tau_4$ and $\tau_5$, the passenger's destination is grid $\#8$ in $\tau_4$ and grid $\#7$ in $\tau_5$, respectively. Thus, the destination probability can be calculated as $P_{dest}(\#1, \# 7) = \dfrac{n_{dest}(\#1, \#7)}{n_{pickup}(\#1)} = \dfrac{1}{2} = 50\%$ and  $P_{dest}(\#1, \#8) = \dfrac{n_{dest}(\#1, \#8)}{n_{pickup}(\#1)} = \dfrac{1}{2} = 50\%$.

In trajectories $\tau_2$ and $\tau_3$, the driver drives to grid $\#2$ to pick up the passenger, and the passenger goes to grid $\#8$. During the trip, the driver has a $P_{match}$ of receiving a new order before she arrives at the destination of the passenger. The order matching probability while on trip can thus be calculated as $P_{match}(\#8) = \dfrac{n_{match}(\#8)}{n_{trips}(\#8)} = \dfrac{1}{2} = 50\%$. 

Based on the probabilities calculated above, an example of state transition is presented in Figure~(\ref{subfig:state_transition_small}). To make the state transition consistent with the five trajectories, we suppose the driver is initially in grid $\#1$ with a status indicator $I$, i.e., the driver is in state $s_0 = (\#0, 0, I)$. If $I=0$, meaning that the driver needs to seek for an e-hailing order, the driver drives into grid $\#1$ and seeks for e-hailing orders in the grid. There are two possible outcomes associated with this case.

\begin{enumerate}
	\item The driver fails to find any e-hailing order in grid $\#1$. The driver will end up in state $s_1 = (\#1, 2, 0)$ and receive a negative reward $-\alpha \times dist(\#0, \#1)$, which is actually the fuel cost.   This outcome happens with probability $p_1 = 1 - P_{order\_match}(\#1) = 1-80\% = 20\%$
	\item The driver successfully finds an e-hailing order in grid $\#1$. For the purpose of demonstration, we assume the driver goes to grid $\#2$ to pick up the passenger, and the destination of the passenger is grid $\#8$. The probability of this outcome is $p_2 = P_{order\_match}(\#1) \times P_{pickup}(\#1, \#2) \times P_{dest}(\#2, \#8)= 80\% \times 50\% \times 100\% = 40\%$. The driver will receive a total reward of $r = fare(\#2, \#8) - \alpha \times [dist(\#0, \#1) + dist(\#1, \#2) + dist(\#2, \#8)]$ by completing this ride.  During the trip, the driver may have a probability $P_{match}(\#8) = 50\%$ of getting a new request before she arrives at the destination of the previous passenger. Hence, there are two possible subbranches from this outcome. 
	\begin{enumerate}
		\item If the driver is matched to a new request before she drops off the previous passenger, then the driver will end up in state $s_2 = (\#8, 7, 1)$. The probability of the occurrence of this subbranch is $p_2 \times P_{match}(\#8) = 40\% \times 50\% = 20\% $.
		\item if the driver fails to be matched to another request while on trip, the driver will then end up in $s_3 = (\#8, 7, 0)$. This subbranch occurs with a probability $p_2 \times (1-P_{match}(\#8)) = 40\% \times (1-50\%) = 20\%$
	\end{enumerate}
\end{enumerate}

For the sake of the completeness of the state transition, the other two subbranches associated with $I=1$ are also displayed in Figure~(\ref{subfig:state_transition_small}). These two subbranches are quite self-explanatory, and thus the detailed discussion will be omitted.

\section{Sequential MDPs for multiple agents} \label{sec:multi_mdp}
Note that the deterministic policy derived is only applicable when there is one agent following the policy. Otherwise there can be local competition among e-hailing drivers since several drivers may be guided into the same grid. We thus need to address the competition among e-hailing drivers if there are multiple idling e-hailing drivers being present in the same region within a short time interval. %Intuitively, the order matching probability for a subsequent driver of going into a grid is supposed to be decreased when some drivers have already been guided into that grid. 
\cite{lin_efficient_2018} proposed a contextual multi-agent reinforcement learning approach in which the multi-agent effect is captured by attenuating the reward through an averaging fashion. \cite{zhou_optimizing_2018} employed a simple discounting factor $\dfrac{1}{n}$ to update the order matching probability when the $(n + 1)^{th}$ taxi is being guided to a road if there are already $n$ taxis going to that road. The discounting factor proposed $\dfrac{1}{n}$ is effective in the sense that it makes the order matching probability smaller for subsequent taxis following the policy. However, the simple discounting factor may underestimate the order matching probability since the effect of the number of orders in each grid was neglected. In other words, except the effect of the number of drivers being guided into a grid, there is an underlying correlation between the decrease in the order matching probability and the number of orders in that grid. Here we use an example to illustrate the existence of the aforementioned correlation. We suppose an e-hailing driver is guided into grid $l$ with an order matching probability 50\%. We consider two extreme scenarios: (1) there was 1 order emerging in grid $l$ and (2) there were infinite orders emerging in grid $l$. After one driver is guided into grid $l$, for a second driver, the order matching probability in grid $l$ is supposed to decrease substantially in the first scenario while almost keeps the same in the second scenario. The rationale underlying this argument is that compared to a grid with a smaller number of orders, a grid with a larger number of orders is capable of accepting more cruising drivers while still maintain a relatively acceptable level of order matching probability.

To incorporate this correlation, we develop a dynamic adjustment strategy. Before formally providing the form of the strategy, we list four intuitive observations: (1) The order matching probability for the first driver being guided into grid $l$ is simply $P_{order\_match}(l)$; (2) The order matching probability for the $n^{th}$ driver being guided into grid $l$ decreases with $n$, meaning that the order matching probability is getting smaller when there are more drivers cruising vacantly in grid $l$; (3) For the $n^{th}$ driver, the order matching probability increases with the number of orders in grid $l$, meaning that a grid with a larger number of orders is able to accept more cruising drivers; (4) Under the extreme scenario where there are infinitely many orders in grid $l$, the order matching probability keeps its level at $P_{order\_match}(l)$ regardless of the number of drivers being guided into grid $l$, as long as it is finite. Based on these four observations, we postulate that the order matching probability of the $n^{th}$ driver in grid $l$ takes the exponential form, i.e.,

%where the order matching probability in a grid is decreased exponentially with the number of drivers being guided into the grid and the effect of the number of orders in the grid is captured by explicitly using the number of orders in the denominator of the coefficient of the exponential function, as shown in Eq~(\ref{eqn:model}).}

\begin{equation}
Pr(l, n) = P_{order\_match}(l) \times e^{-\dfrac{\beta}{\# orders(l)}\times (n - 1)}
\label{eqn:model}
\end{equation}
where $\#orders(l)$ is the number of orders in grid $l$ and $\beta$ is a parameter to be determined.

To calibrate the strategy for the $n^{th}$ driver, the order matching probability $Pr(l,n)$ is required. Note that $Pr(l,1) = P_{order\_match}(l)$, then Equation~(\ref{eqn:model}) can be written as
\begin{equation}
\dfrac{Pr(l,n)}{Pr(l,1)} = e^{-\dfrac{\beta}{\# orders(l)}\times (n - 1)}
\label{eqn:model2}
\end{equation}

To utilize linear regression, we further take the logarithm and then the reciprocal of both sides of Equation~(\ref{eqn:model2}), and thus Equation~(\ref{eqn:model2}) can be rewritten as
\begin{equation}
\dfrac{1}{log\dfrac{Pr(l,n)}{Pr(l,1)}} = - \dfrac{1}{\beta \times (n-1)} \times \#orders(l)
\label{eqn:model3}
\end{equation}

Denoting the left hand side of Equation~(\ref{eqn:model3}) as $prob_n(l)$, the purpose of the calibration is simply to verify the existence of the linear correlation between the two variables $\#orders$ and $prob_n$ and determine the parameter $\beta$. $\#orders(l)$ is simply the number of historical orders in grid $l$. The order matching probability for the $n^{th}$ driver entering grid $l$ can be calculated as follows. We use 18 10-minute intervals to split the 3-hour morning peak used in this study. For each interval, the number of orders within the interval is counted, and it is a success if the counted number is not less than $n$. The probability of success across 18 intervals is $Pr(l,n)$. Calculating $prob_n(l)$ and $\#orders(l)$ for all grids, samples of $prob_n(l)$ and $\#orders(l)$ are obtained.
%Samples of the two variables can be obtained through Algorithm~(\ref{alg:prob_calculation}).
\iffalse
\begin{algorithm}[H]
	\caption{Order matching probability calculation for the $n^{th}$ driver, $Pr(l,n)$}
	\label{alg:prob_calculation}
	\begin{algorithmic}[1]
		\State Initialize an empty list $\#orders$ and an empty list $prob_n$ 
		\State Initialize time intervals $TI$ (in this work, the morning peak is from 7 AM to 10 AM, we use 10-minute intervals to split the 3-hour morning peak into 18 time intervals)
		\For{\texttt{$l \in L$}}
		\State Count the number of orders in grid $l$ and append it to the end of the list $\#orders$
		\State Initialize an empty list $cnt_n$
		\For{\texttt{$ti \in TI$}}
		\State Count the number of drivers successfully gets an order match in grid $l$, denoted as $num\_driver$
		\If {$num\_driver \geq n$}
		\State Append 1 to the end of the list $cnt_n$, indicating a successful order match for the $n^{th}$ driver
		\Else
		\State Append 0 to the end of the list $cnt_n$, indicating an unsuccessful order match for the $n^{th}$ driver
		\EndIf
		\EndFor
		\State Calculate $Pr(l,n) = \dfrac{sum(cnt_n)}{len(cnt_n)}$
		\State Append $\dfrac{1}{log\dfrac{Pr(l,n)}{Pr(l,1)}}$ to the end of the list $prob_n$ 
		\EndFor
		\State Return the lists $prob_n$ and $\#orders$
	\end{algorithmic}
\end{algorithm}
\fi

\begin{figure}[H]
	\centering
	\includegraphics[width=0.7\linewidth,height=\textheight,keepaspectratio]{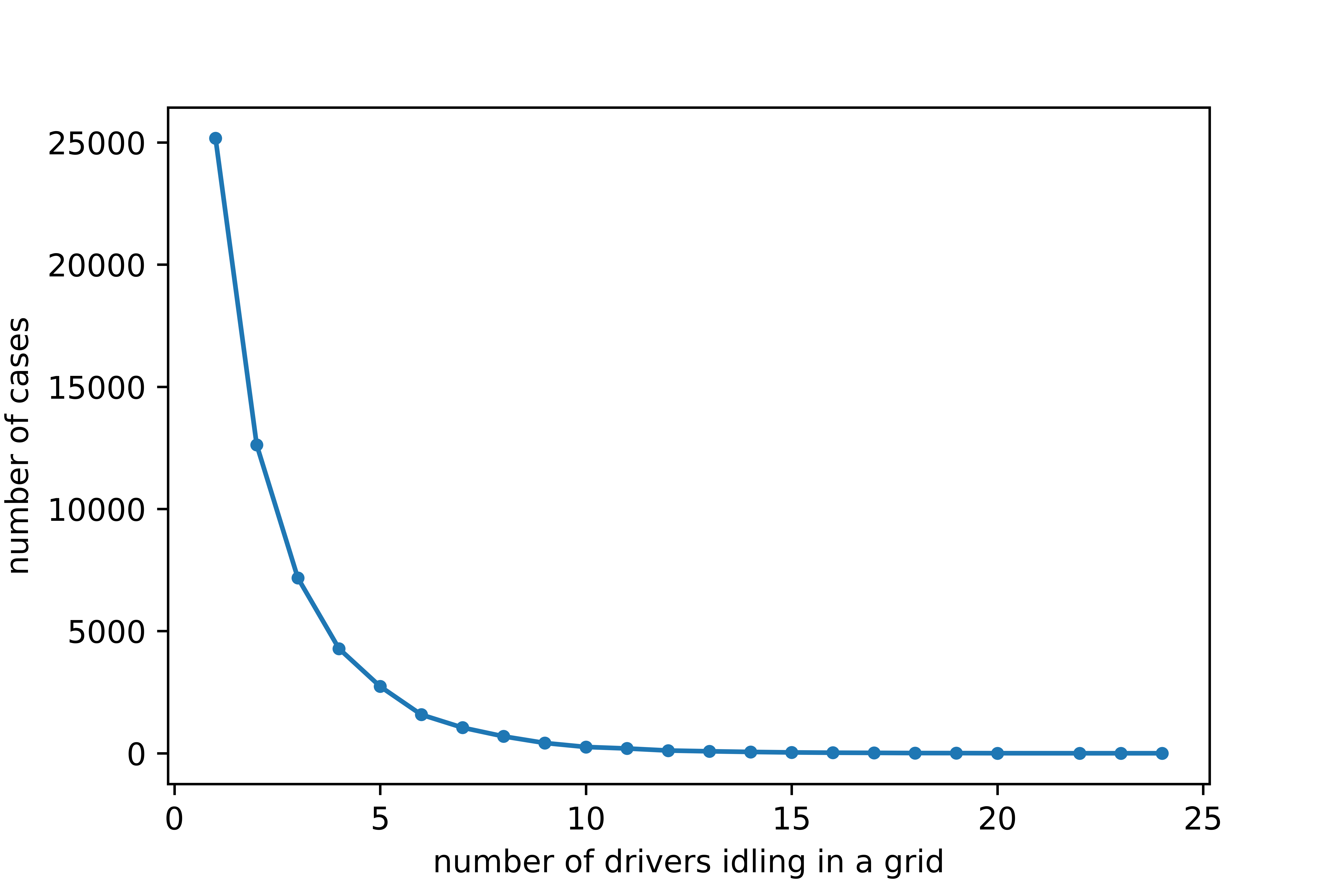}
	\centering 
	\caption{Distribution of number of cases}
	\label{fig:case}
\end{figure}

Ideally, the calibration can be run for any integer value of $n$. However, due to the relatively small size of the grid and the finite number of idling drivers, $n$ cannot be taken as a very large number. To get an appropriate upper bound of $n$ in the calibration, we simply count the number of cases in which there are $n$ drivers idling in a grid within a time interval. The distribution of the number of cases versus the number of idling drivers in a grid is presented in Figure~(\ref{fig:case}). For example, there are more than 25,000 cases in which there are only one idling driver in a grid within a 10-minute time interval. The trend is decreasing, indicating that the number of cases for more drivers idling in a grid within a time interval is less, which is as expected. To obtain statistically meaningful results in the calibration, we need as many samples as possible. Thus, here we choose to do the calibration for $n$ up to 4.

\begin{figure}[H]
	\centering
	\includegraphics[width=0.7\linewidth,height=\textheight,keepaspectratio]{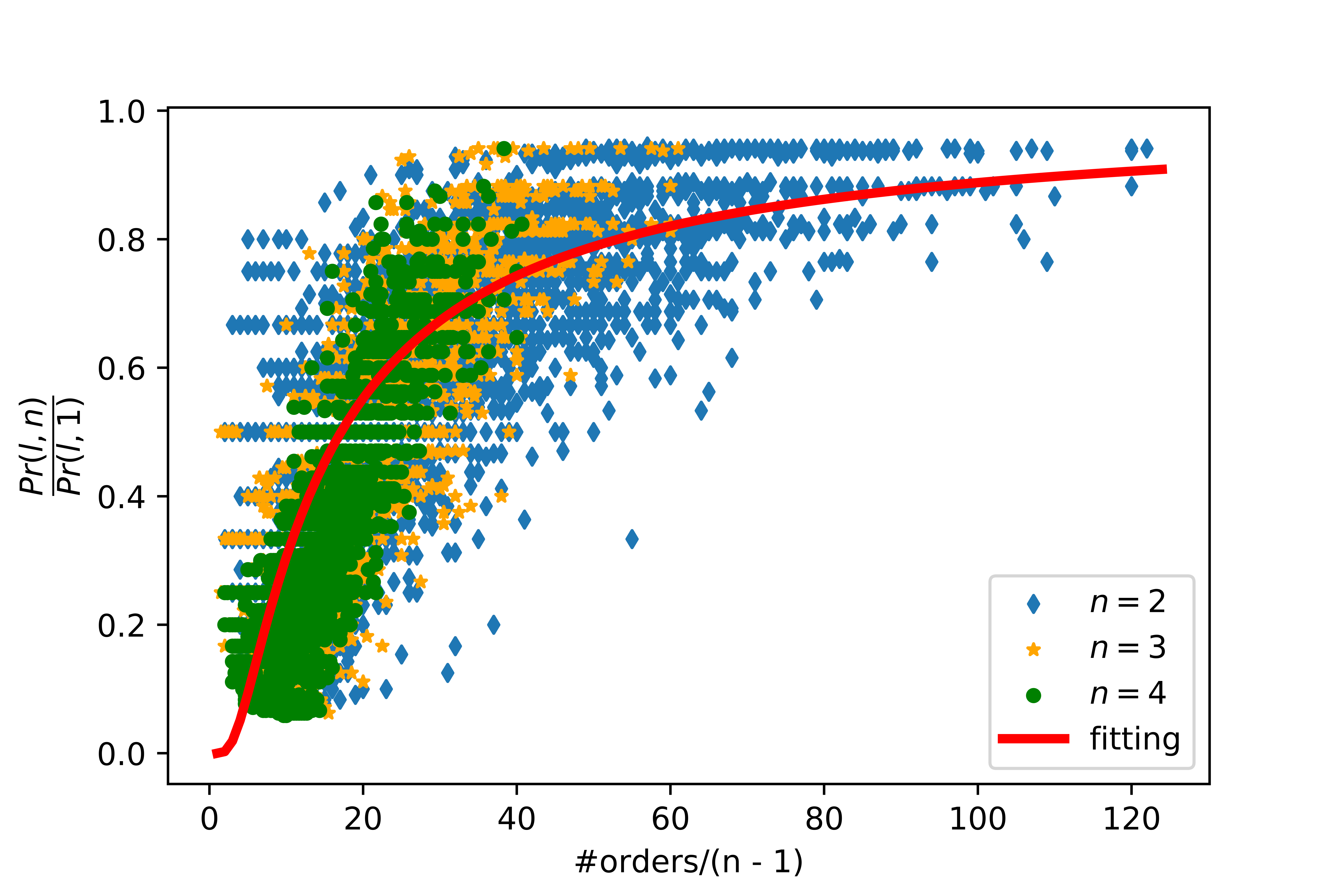}
	\centering 
	\caption{Exponential fitting}
	\label{fig:fitting}
\end{figure}

We obtain three lists of $prob_n$ and three lists of $\#orders$ for $n = 2, 3,$ and $4$, although the three lists $\#orders$ for different $n$ are identical. Noticing that here we only have one parameter $\beta$, we concatenate the three lists of $prob_n$ simply into a long list $prob = [prob_2, prob_3, prob_4]$ and concatenate the three lists of $\#orders$ as a long list $\#orders\_long = [\#orders, \dfrac{\#orders}{2}, \dfrac{\#orders}{3}]$. Then we run a linear regression through origin of the list $prob$ over $orders\_long$. Here we choose the linear regression through origin because the model is naturally linear without intercept, as shown in Equation~(\ref{eqn:model3}), stemming from the requirement that the order matching probability for the first driver in a grid $l$ is $Pr(l,1) = P_{order\_match(l)}$. %and (2) we ran a linear regression with intercept of the list $prob$ over $orders\_long$, and the intercept was found to be roughly 4\% of the mean of the response variable, indicating that the intercept is quite small and dropping the intercept from the linear regression is acceptable. 
The regression result suggests the slope $-\dfrac{1}{\beta} = -0.0842$ with p-value less than $0.001$, indicating that the slope is significant. The R-square value is determined as 65\%, indicating the fitted linear model is able to explain 65\% of the variability and thus the adoption of a linear regression is reasonable. To visually show the goodness of the fitting, we substitute the determined value of $\beta$ into Equation~(\ref{eqn:model}) and plot the fitted curve together with the data for $n = 2, 3$, and $4$ in Figure~(\ref{fig:fitting}).

Based on the calibrated dynamic adjustment strategy, we can now build sequential MDPs for multiple agents and then derive one optimal policy for each agent sequentially. Recall that the dynamic adjustment strategy actually attenuates the order matching probability for the $n^{th}$ driver to be guided into grid $l$ when there are $n-1$ drivers idling in grid $l$. Thus, the general MDP framework proposed in the previous section can directly be applied to the $n^{th}$ driver with a table of relatively lower order matching probabilities. In other words, there is one MDP model for each agent when multiple agents coexist, and the main distinction among these MDP models is the order matching probability (i.e., order matching probabilities of the $n^{th}$ driver are smaller than that of the $(n-1)^{th}$ driver). The optimal policy for each agent can then be obtained by solving her MDP model.

\section{Case study} \label{sec:data}

Nowadays, GPS-enabled devices are ubiquitously used in different types of applications. Especially for drivers, GPS devices can not only help them navigate but also record the real-time location and speed of the vehicle. Hence, a large amount of GPS trajectories, representing sequences of time-stamped geographical points, have been collected and can be a valuable asset for people to understand and tackle real-world traffic problems \citep{di2010hybrid,di2015indifferenceband,shou_similarity_2018}. 

In this research we use large-scale real-world historical GPS traces collected in Beijing in the morning peak, i.e., (7 AM, 10 AM), of the first 3 weekdays in November 2017 by Didi Chuxing, China's leading ride-hailing company. The dataset contains recorded GPS traces of 44,160 e-hailing vehicles and 158,784 e-hailing orders. Whenever an e-hailing driver is online, i.e., she has turned on the e-hailing application, one data point is sampled every 3 seconds. Each data point contains the vehicle's location (i.e., longitude and latitude), current timestamp, the fare of the current occupied trip, if applicable, and a status indicator to record the taxi's current operating state, which includes idle, after matching before pick-up, waiting at the pick-up location, and on trip. %Considering the cyclic pattern of the transportation system, we use three 3-hour time intervals, namely, (7 AM, 10 AM), (12 PM, 3 PM), and (5 PM, 8 PM) to capture the morning peak, off peak, and evening peak. 

Based on the characteristics of the spatial distribution of the passenger pick-up spots and passengers' destinations, we construct a bounding box within the six ring road to cover the city area in which 90\% of the e-hailing orders are preserved. The e-hailing orders which fall outside the bounding box will be disregarded. A hexagonal grid world setup is then adopted, and the city area within the bounding box is split into 6,421 hexagonal grids with the length of the diagonal of a hexagon of approximately 700 meters.

Now we use the IRL technique to uncover the general reward function for all drivers, i.e., to derive the parameter $\alpha$. The observed policy $\pi^O(s)$ in each state $s$ is simply the most frequently taken action by all drivers in that state. Applying the aforementioned IRL technique with two known reward functions (i.e., fare $\phi_1(s,s')$ and traveling distance $-\phi_2(s,s')$) and setting $\alpha_1 = 1$ (i.e., assuming the driver will earn all the fare), we obtain $\alpha = \alpha_2 = 0.64$. %Thus, the parameter $\alpha = \dfrac{\alpha_2}{\alpha_1} = 0.64$. 
In other words, the coefficient of fuel consumption and other operating costs per unit distance is $0.64$ Chinese Yuan. The general reward function applicable to all drivers is $R(s,s') = \phi_1(s,s') - 0.64\times \phi_2(s,s')$. 

%\subsection{Successful drivers}
%\subsection{Statistics for e-hailing drivers}

Different from other public transportation modes, including buses and subways, which are operated according to a fixed schedule and a predefined route, e-hailing drivers are free to choose their own actions after completing a ride, resulting in a discrepancy in drivers' income. In general, an experienced driver is familiar with the city where she operates the vehicle and is usually aware of where to go after completing a ride in order to get the next request as soon as possible. For example, an experienced driver knows where and when passenger demands will be high near some attractions, hotels, or transportation terminals. In practice, however, there is no guarantee that all drivers are experienced and have a good judgment of where to go next. In particular, the popularity of e-hailing applications has lowered the entry barrier of becoming a driver and brought a tons of rookie into the e-hailing market. %Thus, we will need to identify who are more successful drivers. 
To gain some basic understanding of the performance of e-hailing drivers, we adopt two metrics, namely, the rate of return and the utilization rate.

\begin{defn}
	(Rate of return) An e-hailing driver's rate of return on one day is defined as the ratio of the driver's net income (i.e., gross income minus the operating cost) to the driver's working time. 
\end{defn}

\begin{defn}
	(Utilization rate) The utilization rate of an e-hailing vehicle is defined as the ratio of the time spent on carrying a passenger to the total operating time of the vehicle.
\end{defn}

\begin{figure}[H]
	\centering 
	\subfloat[Distribution of the rate of return]{\includegraphics[scale=.5]{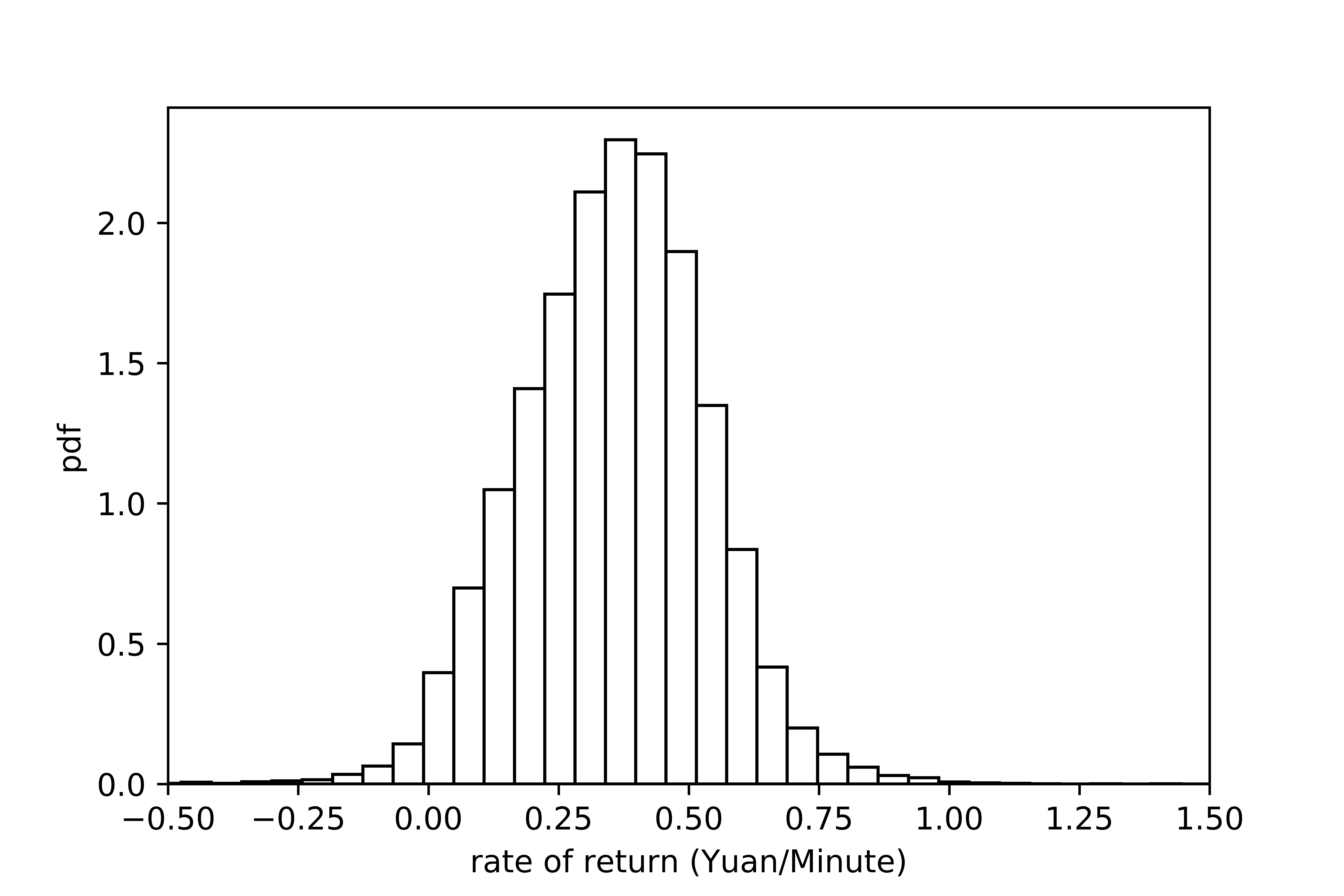}\label{subfig:rate}}~
	\subfloat[Distribution of the utilization rate]{\includegraphics[scale=.5]{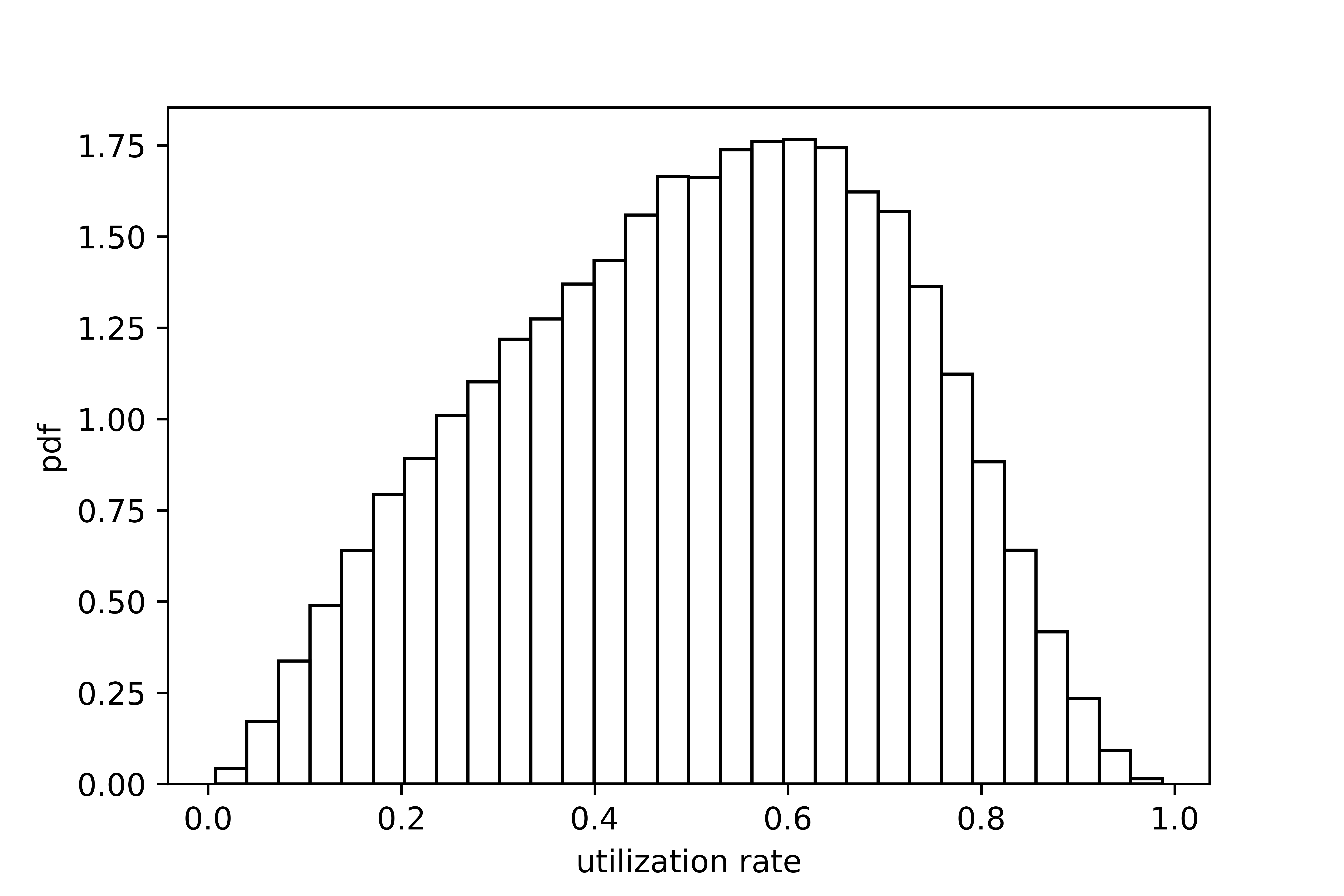}\label{subfig:utilization}} 
	\caption{Distribution of the rate of return and utilization rate from the field data}
	\label{fig:rate_utilization} 
\end{figure}

The rate of return measures a driver's earning ability per unit time. Thus, compared with the total income, which can be largely influenced by a driver's working time, the rate of return is a better metric to be used to measure the performance of drivers. 
Figure~(\ref{subfig:rate}) presents the probability density function (pdf) of the rate of return of all drivers across morning peaks on the first three weekdays in November 2017. The average rate of return is 0.36 (Yuan/minute). About 80\% drivers have a rate of return fall within the range 0.11 to 0.60 (Yuan/minute). Top 10\% drivers can reach a rate of return of 0.60 (Yuan/minute) and higher, while the bottom 10\% have a rate of return below 0.11 (Yuan/minute). %In this research, we simply take the top 10\% drivers as successful drivers. Note that one driver can be among the top 10\% today but may be among the bottom 10\% tomorrow due to the day-to-day variation in drivers' daily income. 
In terms of the utilization rate, real drivers can on average reach 0.51. %while the top 10\% driver can on average reach 0.72 and higher. %The top 10\% drivers we will use in the subsequent analysis is based on their overall behavior for the whole month. 

%\subsection{Results} 

%The top 10\% drivers' 1-month data was then used to train MDP models with a temperature parameter $\beta=100$. 
All drivers' data was then used to train the MDP model. We will then qualitatively examine the optimal policy derived from our MDP model and conduct numerical experiments to evaluate the effectiveness of the policy. 

\subsection{Policy}

\begin{figure}[H]
	\centering
	\includegraphics[width=0.9\linewidth,height=\textheight,keepaspectratio]{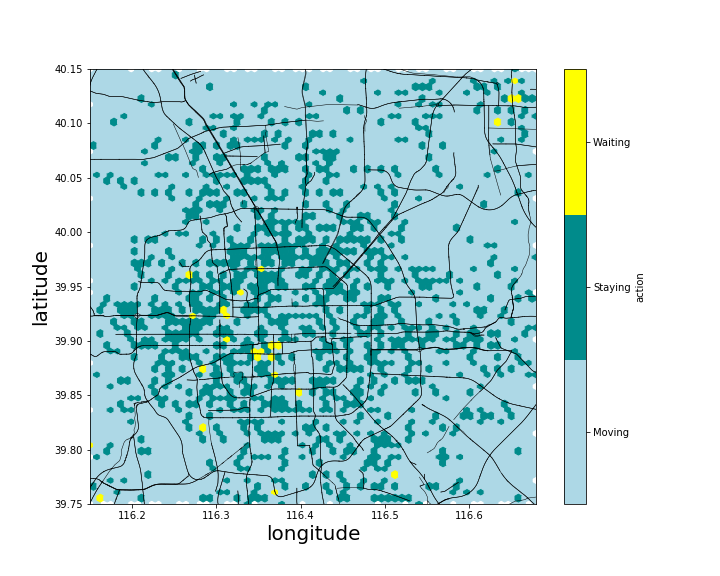}
	\centering 
	\caption{Optimal policy (at the beginning of the morning peak)}
	\label{fig:optimal_policy_overall}
\end{figure}

Figure~(\ref{fig:optimal_policy_overall}) presents the optimal policy %(only the action with the largest probability is taken for the purpose of visualization) 
at the beginning of the morning peak. Light blue color suggests the driver in the current grid to move into one of its neighboring grids to seek for the next potential e-hailing order. The color dark green in a grid stands for staying, meaning that the optimal policy is to stay and cruise within the current grid. The color yellow in a grid means waiting, indicating that the optimal policy for the driver to follow is simply to wait in the current grid. %Some grids are empty because either there is no road in the grid or there is no visitation of any driver to the grid. 
It can be seen that in many grids within the city area (i.e., around the center part of the figure), the optimal policy suggests a driver to stay after completing a ride. In suburban areas, optimal policy usually suggests drivers to move around to some grids with a high probability of receiving a request. Also, there are several places where the probability of receiving a request while waiting is quite high, and the optimal policy advices drivers to wait in these places. %, which is not intuitive in the traditional taxi market. In road-hailing taxi market, except some transportation terminals and some CBD areas where the demand is very high, drivers are supposed to cruise around to find the next passenger in stead of waiting at one spot. The main reason is that for traditional drivers, they have to cruise around to get themselves exposed to more taxi requests. While in e-hailing taxi market, drivers can receive and accept a potential request which is sent from several blocks or streets away. Hence, it is reasonable that our optimal policy suggests drivers to wait within the city area. Outside of fourth ring road area, the optimal policy refers drivers to some local areas with a high probability of receiving a request. 

\iffalse
\begin{figure}[H]
	\centering 
	\subfloat[Morning peak]{\includegraphics[scale=.35]{./figures/p_find_morning}\label{subfig:p_find_morning_2d}} ~
	\subfloat[Off peak ]{\includegraphics[scale=.35]{./figures/p_find_off}\label{subfig:p_find_off_2d}}
	\\
	\subfloat[Evening peak]{\includegraphics[scale=.35]{./figures/p_find_evening}\label{subfig:p_find_evening_2d}}
	\caption{Distributions of the order matching probability $ P_{order\_match}$ during three time intervals}
	%\label{fig:p_finds} 
\end{figure}
\fi

\begin{figure}[H]
	\centering
	\includegraphics[width=0.9\linewidth,height=\textheight,keepaspectratio]{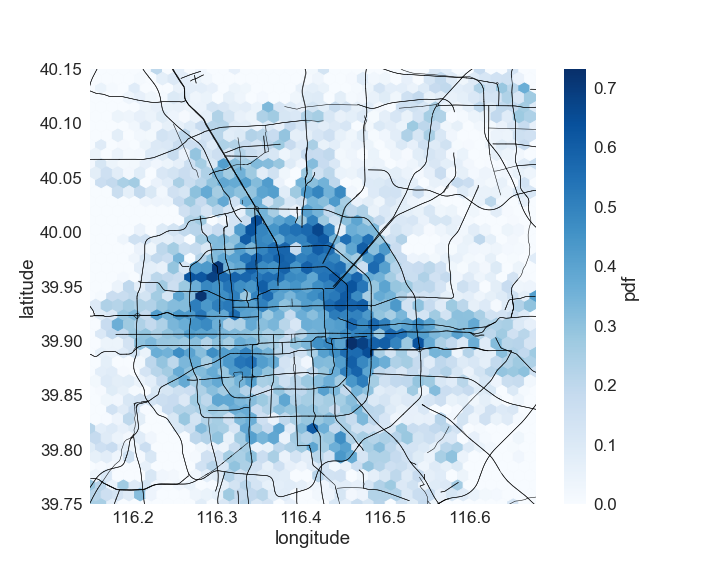}
	\centering 
	\caption{Distribution of the order matching probability $ P_{order\_match}$}
	\label{fig:p_finds}
\end{figure}

Figure~(\ref{fig:p_finds}) plots the distribution of the order matching probability. It can be seen that the majority of the color dark blue, indicating a higher order matching probability, is distributed within the fourth ring road, meaning that the order matching probability is higher in the city area. Furthermore, the distribution of the order matching probability agrees well with the distribution of the optimal policy if we compare Figure~(\ref{fig:p_finds}) with Figure~(\ref{fig:optimal_policy_overall}). The dark green places, indicating staying, in Figure~(\ref{fig:optimal_policy_overall}) generally overlap with blue or dark blue locations in Figure~(\ref{fig:p_finds}), indicating a relatively high order matching probability. Also, when a driver is in a grid with a low order matching probability (e.g., in the nearly white area in Figure~(\ref{fig:p_finds})), the driver needs to move around (as shown by the light blue in Figure~(\ref{fig:optimal_policy_overall})) to enter a grid with a higher order matching probability. This overlapping essentially indicates that grids with a higher order matching probability is more preferable by the agent, compared with a grid with a lower order matching probability.

%During the morning peak, the distribution of the order matching probability is overall in color dark blue, indicating that the overall order matching probability of the morning peak is higher than that of other two time periods. Further, the distribution of the color dark blue is more sparsely distributed over the region of interest during the morning. While during the evening peak, the distribution of the color dark blue is concentrated within the city area, indicating that many demands occurred in the evening peak in the city area, which is as expected since many people will call for an e-hailing vehicle when they get off from work. 

\subsection{Model Evaluation} 
To evaluate the effectiveness of the optimal policy derived from the MDP model, %a Monte Carlo simulation is performed. 
the performance of an agent under the guidance of the optimal policy is compared with that of an agent following one baseline heuristic, i.e., the local hotspot strategy. %The policy random walk essentially suggests the agent to randomly take an action from the allowable actions at the current state. 
The local hotspot strategy essentially suggests the agent to move into grids with a higher demand sequentially and is found to perform the best among three baseline heuristics, namely, random walk, global hotspot, and local hotspot \citep{yu_markov_2019}. 

To obtain the performance of an agent according to different policies, a Monte Carlo simulation is conducted. The basic idea of the simulation is to randomly place an agent in one grid at the beginning, and let the agent move around according to the chosen policy. The environment is determined by the parameters extracted in Section~(\ref{subsec:parameters}). 
In particular, every time when the agent is in a grid $l_a$, we sample a probability of finding a request from a binomial distribution with a success probability $p_{order\_match}(l_a)$. We then sample the pick-up grid and drop-off grid from a multinomial distribution determined by the probabilities $p_{pickup}$ and $p_{dest}$, respectively. The driving time and driving distance can be simply obtained from $t_{drive}$ and $d_{drive}$ after the pickup spot and destination have been determined. 

The simulation is run for millions of times to obtain robust results. We first adopt two metrics, namely, rate of return and the utilization rate of the vehicle to compare the performance of the agent under the guidance of different policies. We then examine the distribution of the number of completed orders, idling time, service time per order, and the profit per unit time of each order.

\begin{figure}[H]
	\centering 
	\subfloat[rate of return (Yuan/Minute)]{\includegraphics[scale=.5]{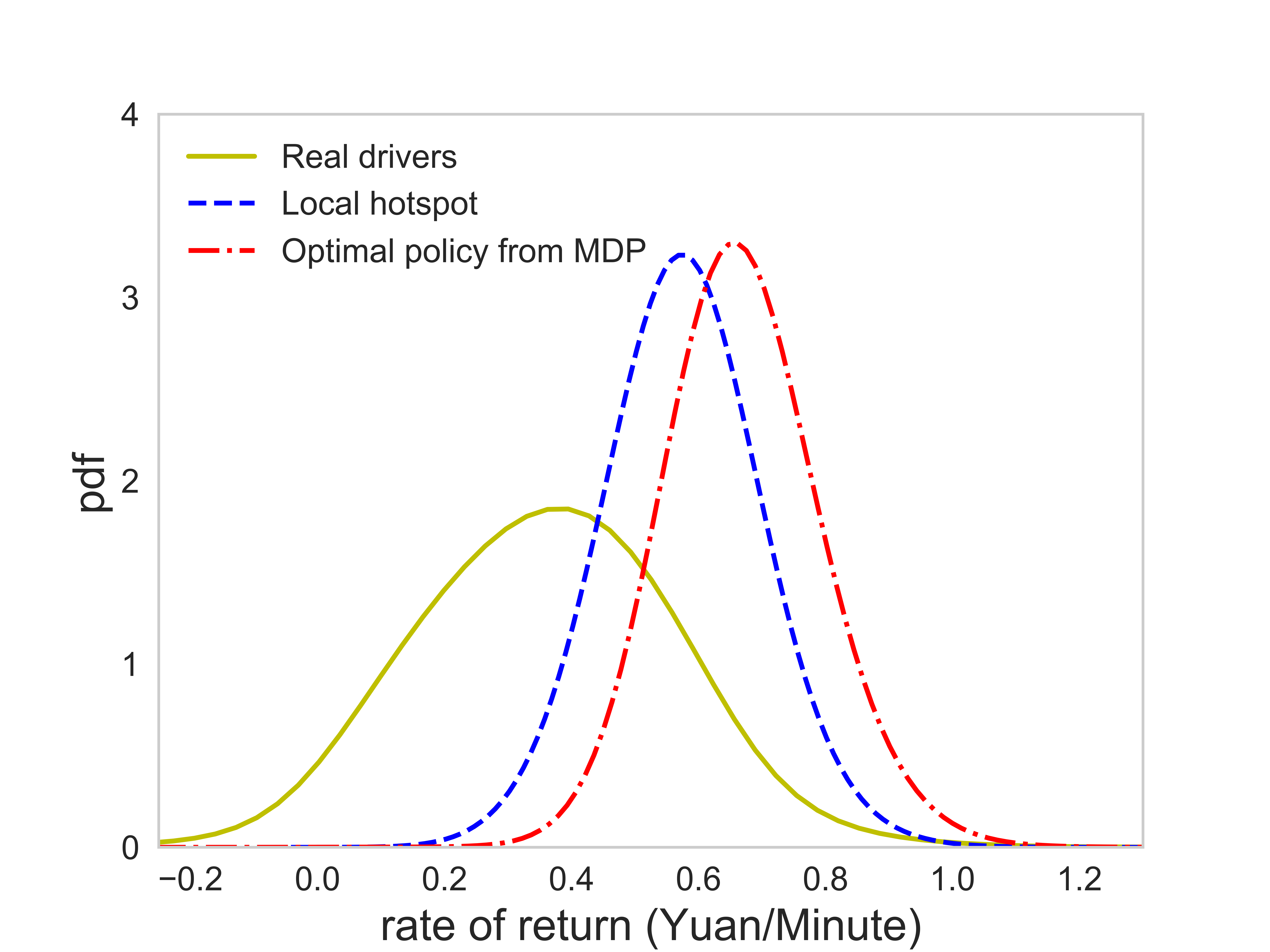}\label{subfig:ror}}~
	\subfloat[utilization rate]{\includegraphics[scale=.5]{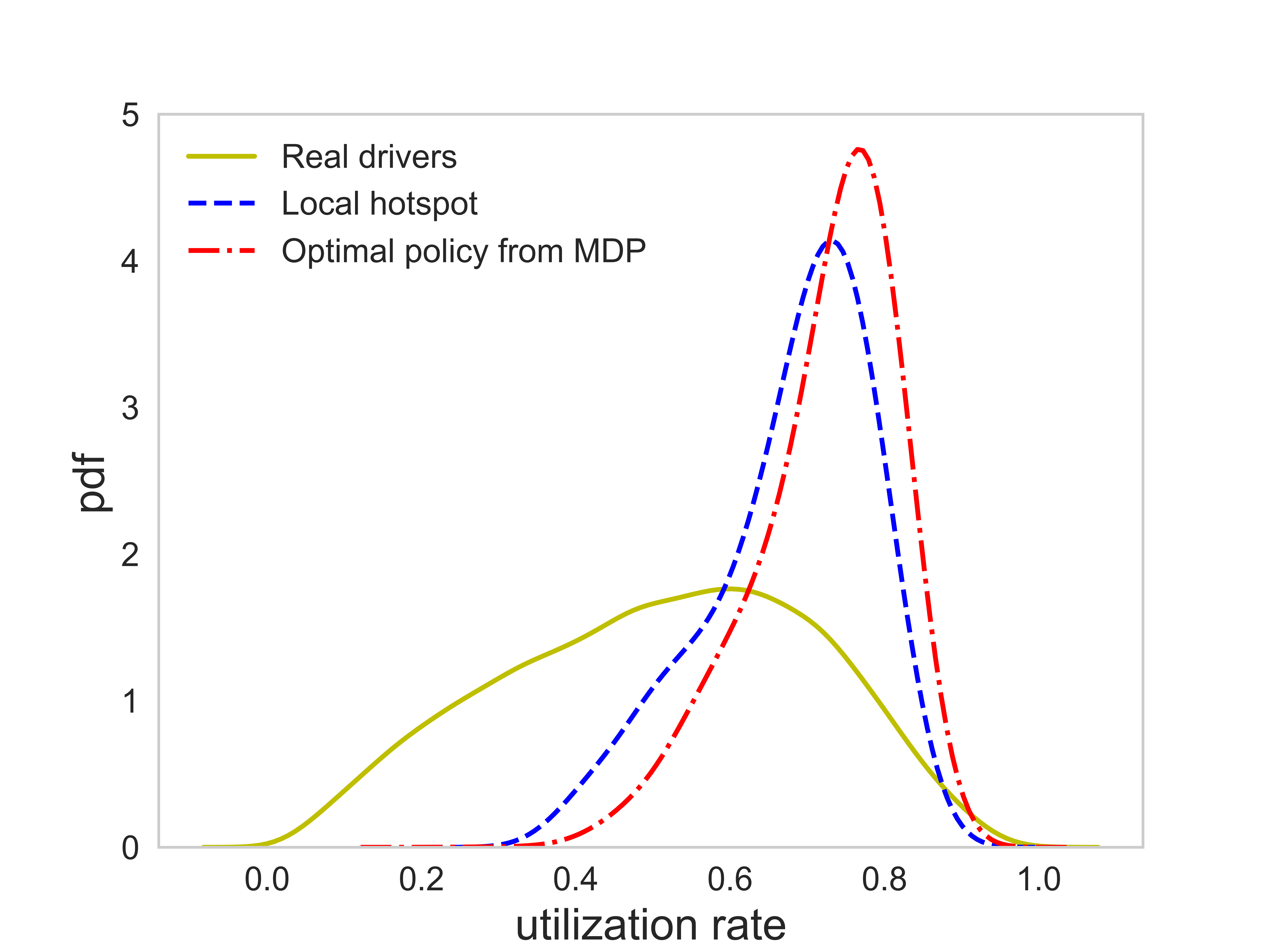}\label{subfig:ur}} 
	\caption{Distribution of the rate of return and the utilization rate of real drivers and the agent}
	\label{fig:compar} 
\end{figure}

Figures~(\ref{fig:compar}) presents the distribution of the rate of return and the utilization rate of the agent following the optimal policy and the local hotspot strategy, respectively. %\textcolor{red}{The performance of real drivers is also included. but not for comparison.?}
It can be seen that the performance of the agent following the optimal policy is on average better than that of the agent following the local hotspot strategy. In terms of the rate of return, the average value that the agent can reach are 0.67 (Yuan/Minute) and 0.57 (Yuan/Minute) under the guidance of the optimal policy and the local hotspot strategy, respectively, meaning that the optimal policy is able to increase the average rate of return by 17.5\% over the local hotspot strategy. The average rate of return of real drivers is 0.36 (Yuan/Minute). %The optimal policy is thus able to achieve a 4.3\% improvement over the top 10\% drivers. 
In terms of the utilization rate, the average value that the agent can reach are 0.72 and 0.67 by following the optimal policy and the local hotspot strategy, respectively, indicating a 7.5\% improvement of the optimal policy over the local hotspot strategy. The average utilization rate of real drivers is around 0.51.

\begin{table}[H]
	\centering
	\caption{Statistics of the comparison between the performance of the agent under different policies}\label{tab:compari}
	\begin{tabular}{|p{4.5 cm}|p{2 cm}|p{4 cm}|p{4 cm}|}
		\hline
		& real drivers & agent following the local hotspot strategy & agent following the optimal policy \\
		\hline
		average rate of return (Yuan/Minute) & 0.36 & 0.57 & 0.67 \\
		\hline
		average utilization rate & 0.51 & 0.67 & 0.72 \\
		\hline
		average number of orders & 6.08 & 8.21 & 8.92 \\
		\hline
		average idling time (Minute) & 47.98 & 27.12 & 22.61 \\
		\hline
		average profit per unit time of each order (Yuan/Minute) & 1.35 & 1.51 & 1.68 \\
		\hline
		average service time per order (Minute) & 16.97 & 14.83 & 14.56 \\
		\hline
	\end{tabular}
\end{table}

\begin{figure}[H]
	\centering 
	\subfloat[Distribution of number of orders]{\includegraphics[scale=.5]{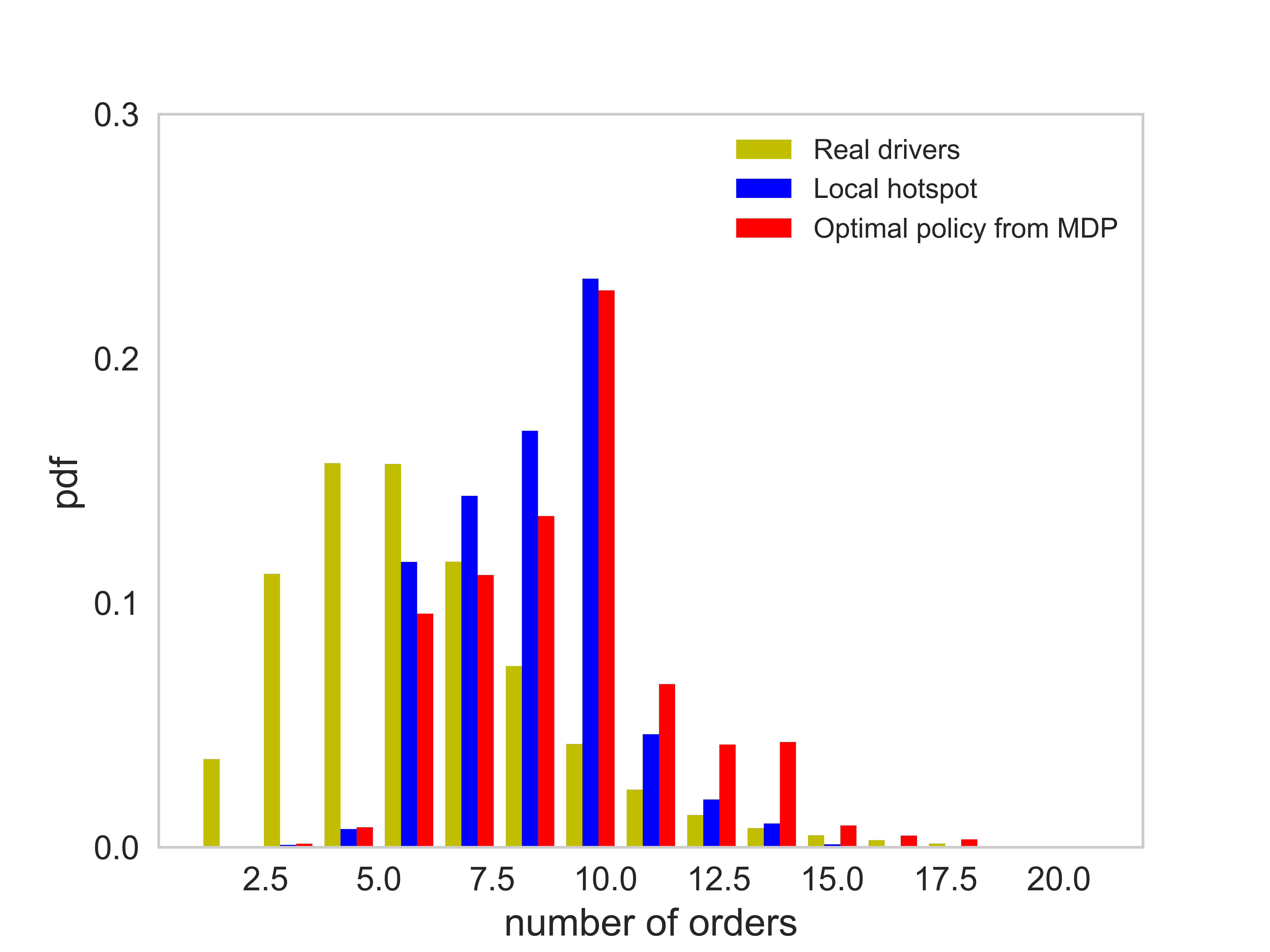}\label{subfig:orders}} ~
	\subfloat[Distribution of idling time ]{\includegraphics[scale=.5]{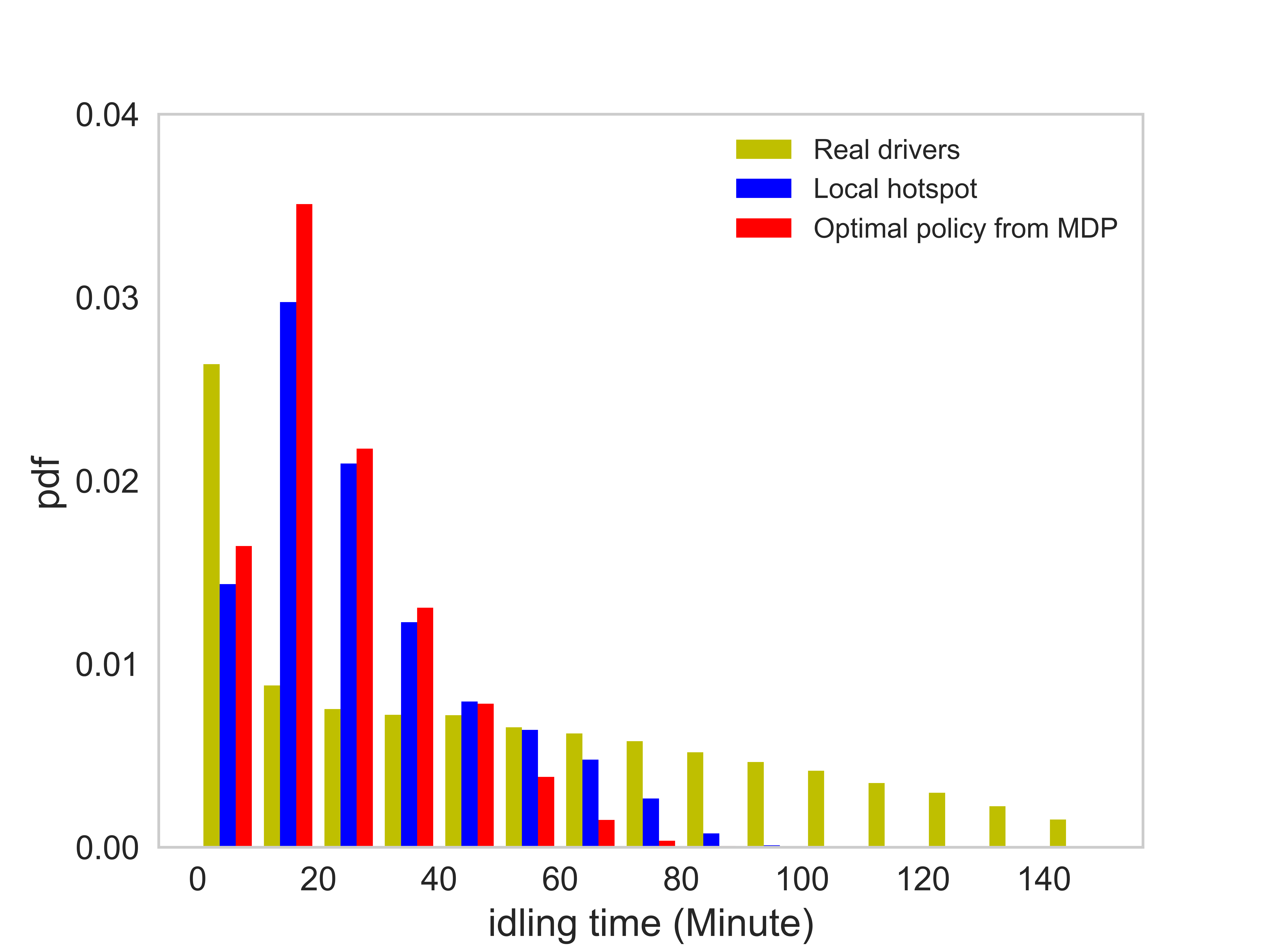}\label{subfig:cells}}
	\\
	\subfloat[Distribution of profit per unit time of each order]{\includegraphics[scale=.5]{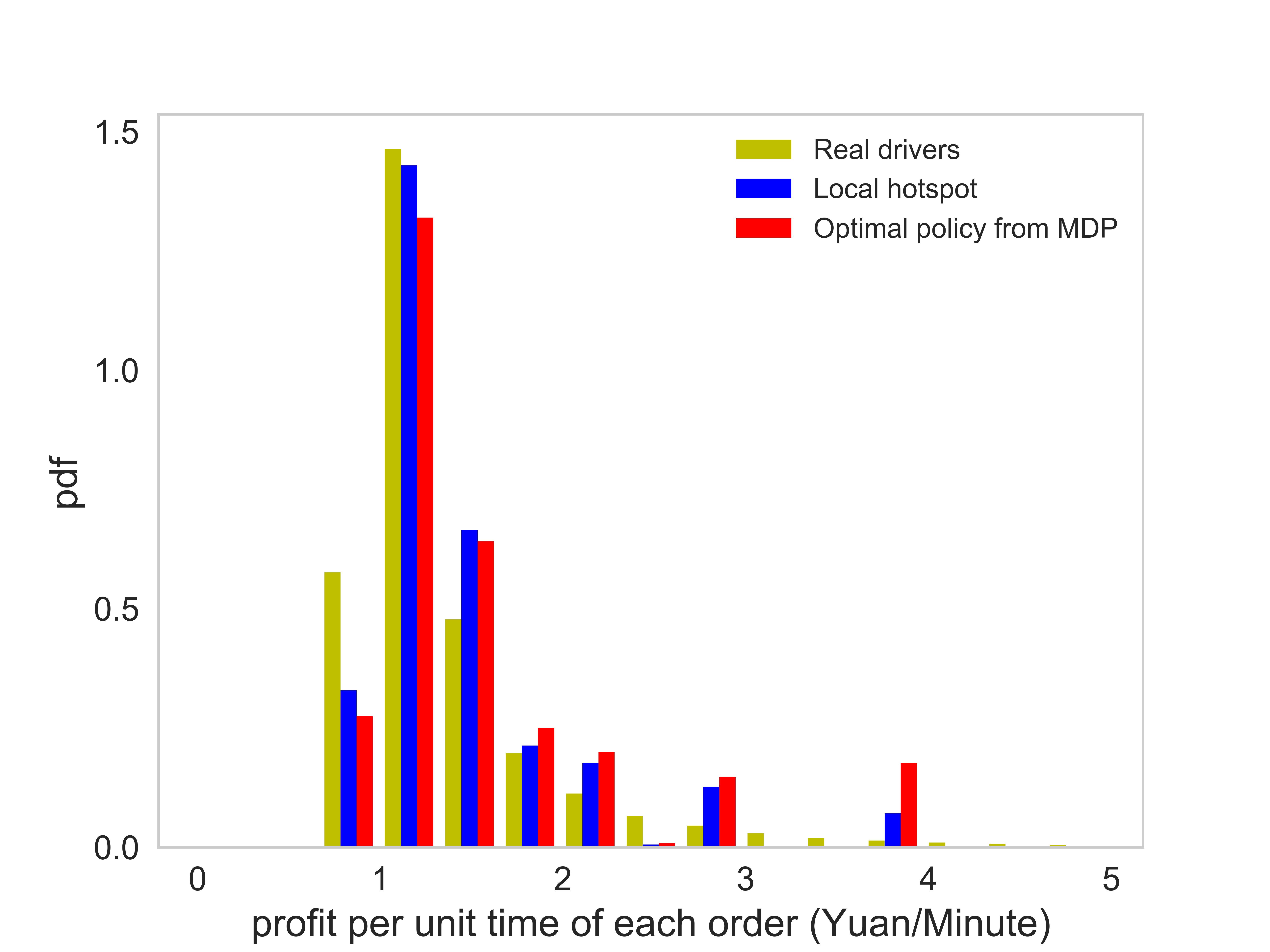}\label{subfig:unit}} ~
	\subfloat[Distribution of service time per order ]{\includegraphics[scale=.5]{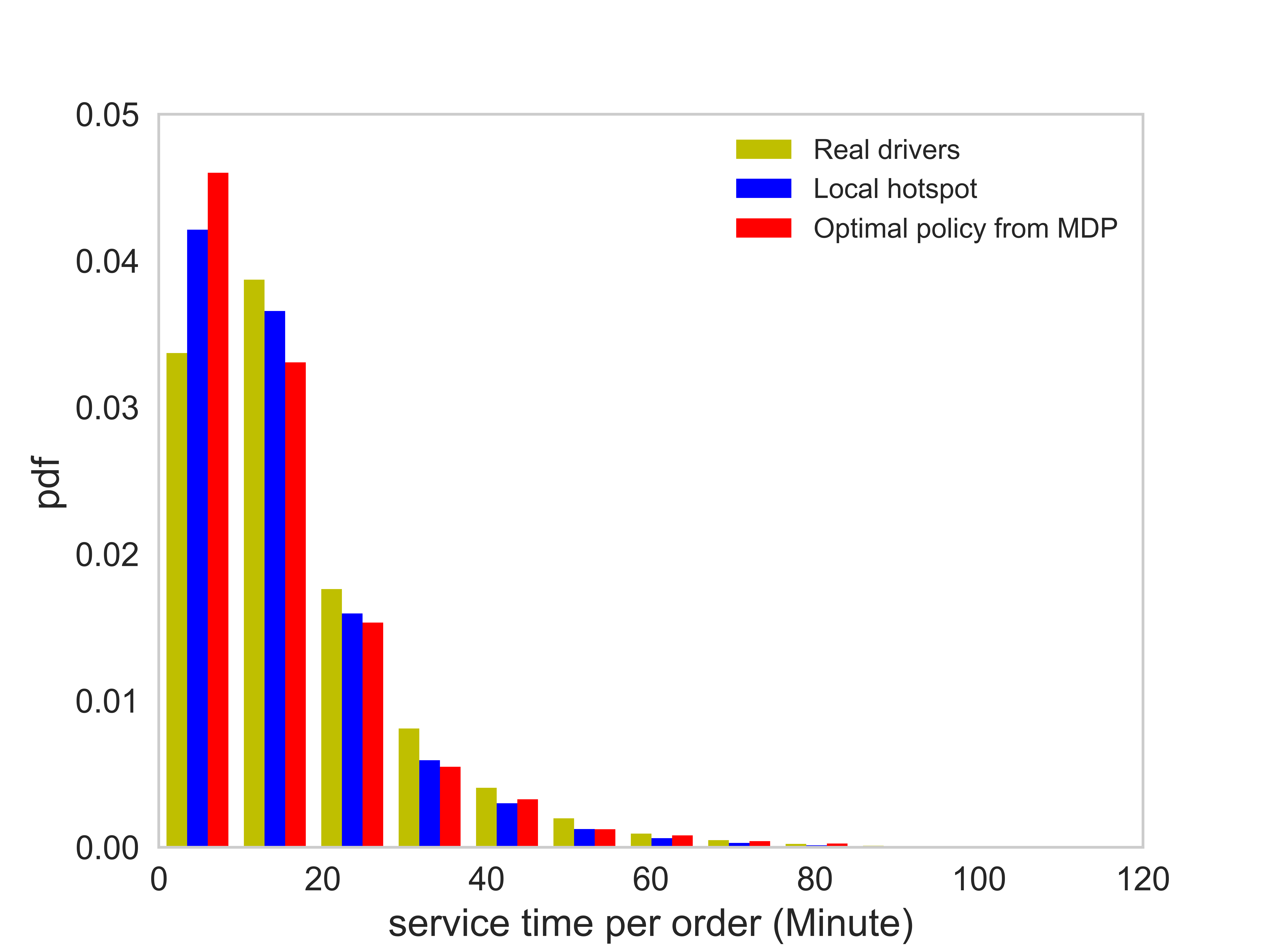}\label{subfig:fares}}
	\caption{Distributions of the number of completed orders, number of idling cells, the service time per order, and the profit per unit time of orders}
	\label{fig:distributions} 
\end{figure}

Figure~(\ref{fig:distributions}) presents distributions of the number of completed orders, idling time, service time per order, and the profit per unit time of each order. The corresponding average value of each case is listed in Table~(\ref{tab:compari}). There are several observations worth mentioning: (1) On average the agent following the optimal policy can achieve a higher number of completed orders than the agent following the local hotspot strategy. (2) The agent following the optimal policy and the local hotspot strategy can achieve a significantly lower idling time, compared with that of real drivers. Thus, both the optimal policy and the local hotspot strategy are able to help agents find requests faster. (3) The distribution of the profit per unit time essentially says that under the optimal policy, the agent is able to find better orders. Here we say an e-hailing order is better when the profit per unit time of the order is higher.
(4) The average service time of orders taken by the agent is shorter than that of real drivers. 

\begin{figure}[H]
	\centering
	\includegraphics[width=0.7\linewidth,height=\textheight,keepaspectratio]{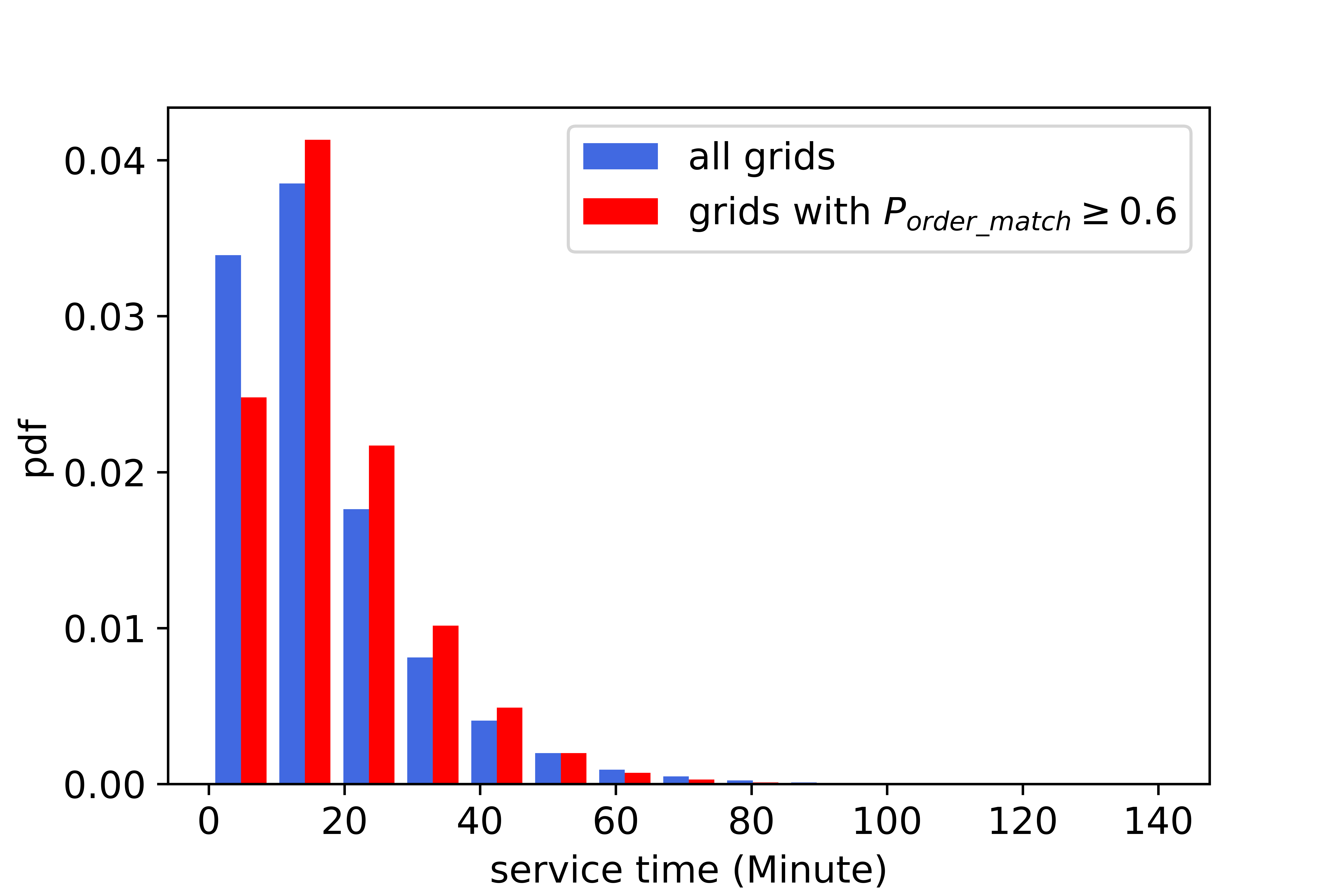}
	\centering 
	\caption{Distribution of service time of orders starting in all grids and in grids with a relatively high order matching probability}
	\label{fig:service_time_grids}
\end{figure}

The aforementioned observation (4), however, does not necessarily indicate that on average a shorter order is more preferable compared with a longer order. The reasons are as follows. As previously stated, for an agent, grids with relatively higher order matching probability is more preferable, compared with grids with a moderate or low order matching probability. Figure~(\ref{fig:service_time_grids}) presents the distribution of service time of orders starting in all grids and in grids with a relatively high order matching probability. The average service time of orders starting in all grids and in grids with a relatively high order matching probability are 17 minutes and 18.34 minutes, respectively. In other words, the passenger requests in grids with a relative higher order matching probability on average has a slightly greater service time. Hence, the average service time per order of the agent is supposed to be slightly higher than that of real drivers, which is not consistent with the observation (4).

\begin{figure}[H]
	\centering
	\includegraphics[width=0.7\linewidth,height=\textheight,keepaspectratio]{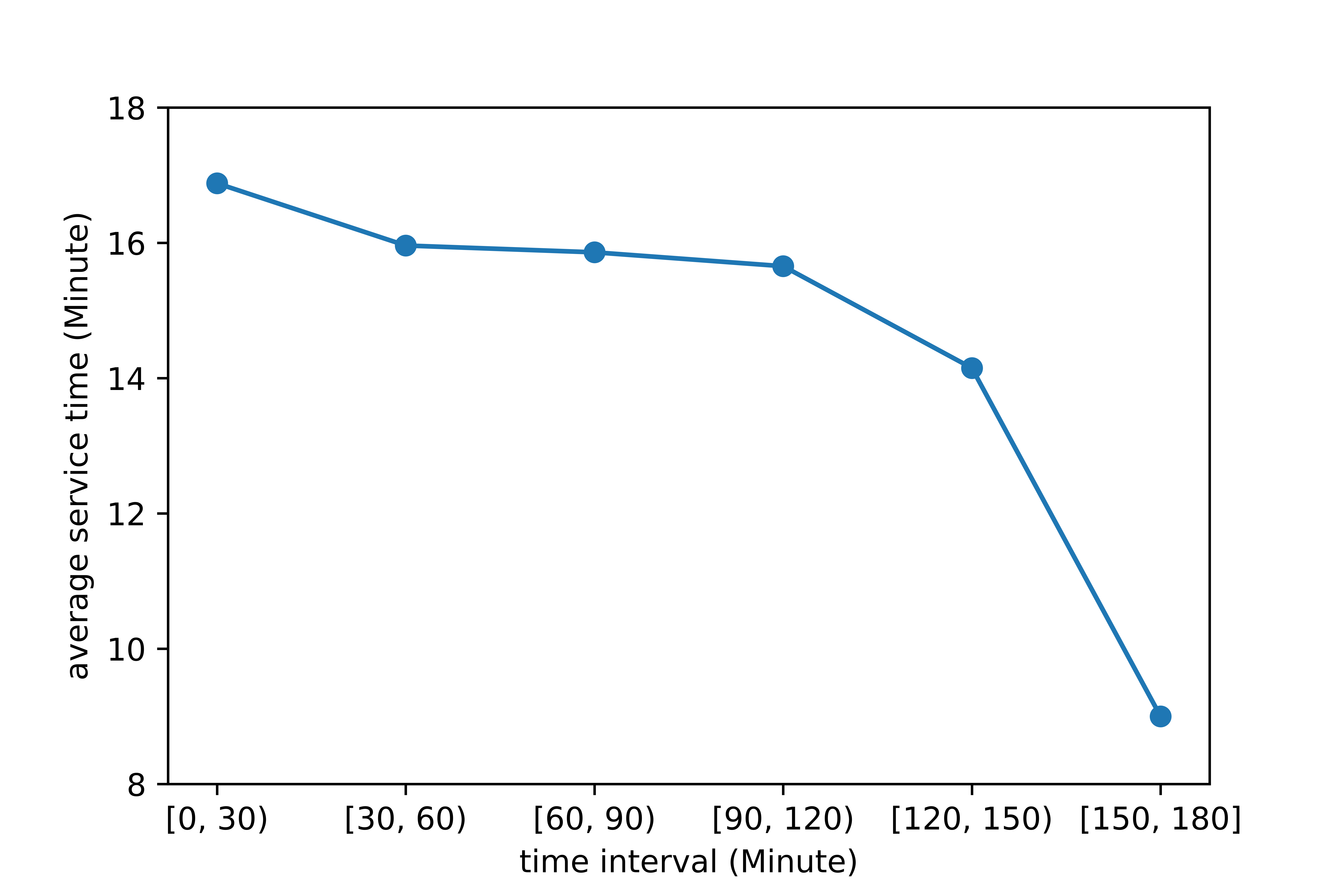}
	\centering 
	\caption{Average service time of orders starting in each subinterval}
	\label{fig:average_fare_interval}
\end{figure}

To explain the inconsistency, we split the 3-hour time interval, i.e., the morning peak, into 6 subintervals of even length, namely, $[0, 30)$, $[30, 60)$, $[60, 90)$, $[90, 120)$, $[120, 150)$, and $[150, 180]$, and then extract the average service time of orders starting in each subinterval, which is shown in Figure~(\ref{fig:average_fare_interval}). The average service time of orders starting in each subinterval decreases as time elapses. In the first subinterval, i.e., around the beginning of the morning peak, the agent actually takes orders with an average service time of 16.88 minutes, which is very close to 16.97 minutes, i.e., the average service time of all orders. This is as expected because the initial position of the agent is randomly chosen, meaning that the distribution of the service time of orders taken by the agent at the beginning of the time interval is supposed to be similar to that of all orders. As time elapses, the average service time of orders taken by the agent decreases because in the simulation we restrict the ending time of an order that an agent can take to be within the 3-hour time interval, and thus the agent to some degree prefers shorter orders, especially when the agent is in the last two subintervals, i.e., around the end of the 3-hour time interval. Low values of the service time of orders in the last two subintervals, i.e., 14.12 minutes and 9.01 minutes largely drag down the overall average service time of orders taken by the agent and caused the aforementioned inconsistency. Here we emphasize that in the simulation, the agent is set to complete the 3-hour time interval, and the restriction imposed by the 3-hour time interval implicitly forces the agent to take relatively shorter orders. All references listed in Table~(\ref{tab:mdp}) except \citep{gao_optimize_2018} used finite horizon time intervals \citep{rong_rich_2016, verma_augmenting_2017, lin_efficient_2018} or finite pickup-dropoff cycles \citep{yu_markov_2019} in MDP modeling. Actually, in reality, an e-hailing driver is free to stop working at any time as long as she has achieved her preset goal, such as a nominal income, a personalized utility function, etc. Thus, an MDP with optimal stopping time modeling is a more realistic and efficient model and is left for future research.

%The increase in the number of completed orders, the decrease in the number of idle cells, and the ability to find better orders are obviously beneficial to the agent. The reason why the agent prefers a shorter order on average is threefold: (1) On average, the earning ability per unit time of the agent with a shorter order is a bit higher than that of the agent with a longer order; (2) Due to the time constraint in the model, i.e., the total duration is 180 minutes, the agent cannot always choose orders with a long duration since the ending time of the order may fall outside the time interval. In other words, it's beneficial to the performance of the agent if the agent can realize the income as soon as possible, especially when the agent is around the end of the time interval; (3) Short orders usually come with short travel distance and thus are able to keep the agent around areas with overall higher order matching probability, which is a plus for the performance of the agent. On the contrary, a longer order may bring the agent from the city center to the suburban area where the demand is relatively low.

%(1) negative slope of the fare per min versus duration (earning ability per unit time) (2) time constraint, realize the income asap (3) short orders will keep the agent around the city center area where the p\_find is high

\subsection{Supply-demand ratio}
\begin{figure}[H]
	\centering 
	\subfloat[Around Xibeiwang (outside the 5th ring road)]{\includegraphics[scale=.74]{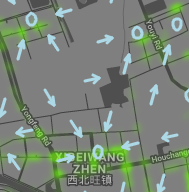}\label{subfig:optimal_policy_consistent}}  \hspace{5mm}
	\subfloat[Around Chaoyang (inside the 3rd ring road) ]{\includegraphics[scale=.77]{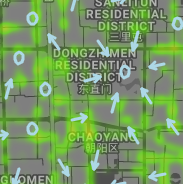}\label{subfig:optimal_policy_inconsistent}}
	\caption{Two zoom-in views of the optimal policy and the distribution of the demand}
	\label{fig:optimal_policy} 
\end{figure}

%Figure~(\ref{subfig:optimal_policy_consistent}) presents a zoom-in view of the optimal policy (shown in red arrows and circles where red arrows suggest the driver to move along the direction denoted by the arrow and the red circle denotes staying or waiting) and the distribution of the demand (shown in heatmap) near Beijing capital international airport. We can see that our optimal policy suggests to stay near the grid which is located at the terminal 2 of the airport. In this grid, the taxi demand is quite high, which can be seen from the background red color, indicating that it is very likely for a driver to receive a request in this grid. As a result, the optimal policy suggests nearby drivers to move into the grid at terminal 2 to take advantage of the high demand in this grid. Also, the demand around terminal 3 is also high, and the optimal policy suggests drivers to stay or wait if they are around the terminal. 

Figure~(\ref{subfig:optimal_policy_consistent}) presents a zoom-in view of the optimal policy (shown in arrows and circles where arrows suggest the driver to move along the direction denoted by the arrow and the circle denotes staying or waiting) and the distribution of the demand (shown in heatmap) near Xibeiwang, an area outside the 5th ring road. We can see that our optimal policy suggests drivers to stay around the grid where the demand is high to take advantage of the locally high demand. Outside the 5th ring road, the overall demand is not very high and the number of idle drivers, indicating the supply, is also not large, resulting in a locally high order matching probability in high demand areas. Thus, a driver can simply take advantage of the high demand to make more profit. 

%Also, notice that several grids away from the terminal 2, the optimal policy suggests drivers to drive to some nearby grids with relatively high demand. The reason can be twofold: either there are some villages near the airport and the driver can easily find a request there or drivers may be matched to a request before dropping of a passenger, resulting in an artificially high demand in some grids near the airport. 

Figure~(\ref{subfig:optimal_policy_inconsistent}) presents a zoom-in view of the optimal policy around Chaoyang district, which is located within the 3rd ring road. Different from the consistency between the policy and the distribution of the demand observed in Figure~(\ref{subfig:optimal_policy_consistent}), now it seems the derived optimal policy and the distribution of the demand is not very consistent. In other words, the optimal policy suggests drivers to move away from the areas with a high demand. The main reason is that the optimal policy is supposed to be consistent with the order matching probability, which captures the supply-demand ratio under the assumption that the number of unmatched order is negligible, instead of the real distribution of the demand. Although the order matching probability is calculated from the distribution of the demand and the number of pass-bys of idle drivers, there may exist some shift or even inconsistency between the order matching probability and the distribution of the demand. For instance, a cell with a very high demand, e.g. 100 order matches, may have a 1,000 pass-bys by idle drivers, resulting in a relatively low order matching probability in the cell, which is 0.1 in this example. Hence, it is not surprising that the derived optimal policy suggests the driver to move away from the cell with a high demand and a low order matching probability. A grid with a high demand may also have a high supply, resulting in a low order matching probability which is not preferable to the agent.

\subsection{Different optimal policies for multiple agents}

\begin{figure}[H]
	\centering 
	\subfloat[Around Dongxiaoying (outside the 5th ring road)]{\includegraphics[scale=.465]{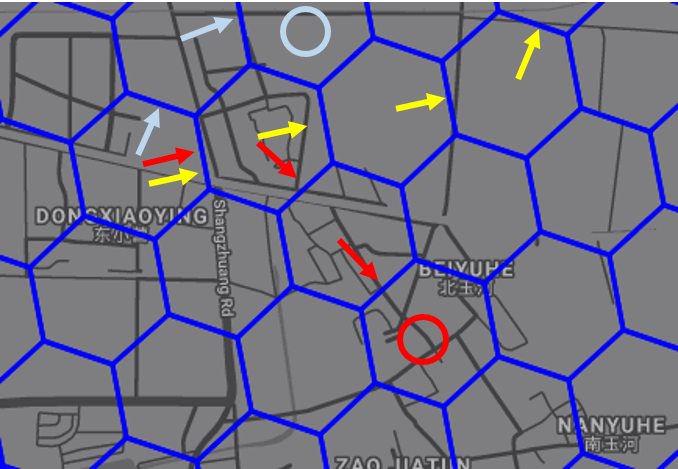}\label{subfig:policy1}}  \hspace{5mm}
	\subfloat[Around Jinrongjie (inside the 2nd ring road) ]{\includegraphics[scale=.47]{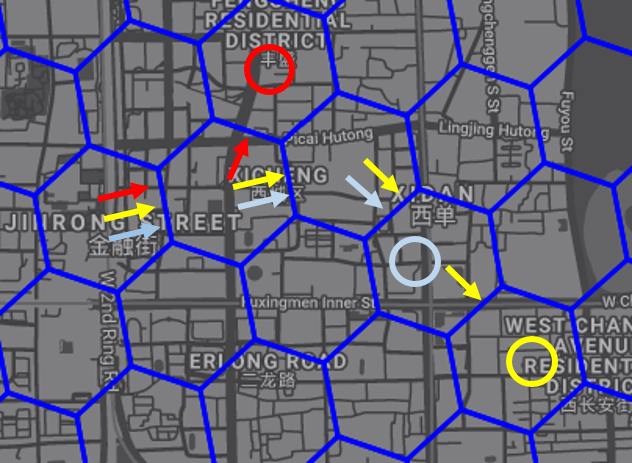}\label{subfig:policy2}}
	\caption{Two zoom-in views of the optimal policy for three agents (red, yellow, and light blue denote the first, second, and third agent, respectively)}
	\label{fig:multi_policy} 
\end{figure}

Now, we verify the effectiveness of the proposed dynamic adjustment strategy of the order matching probability by recommending the next several optimal actions to take for three agents who are currently in the same grid. We randomly selected two places, namely, one place around Dongxiaoying (outside the 5th ring road, thus in suburban area) and one place around Jinrongjie (inside the 2nd ring road, thus in city area).

Figure~(\ref{fig:multi_policy}) presents the recommended optimal actions to take for three agents. In both cases, three agents are being recommended at the same time in the same grid. The results shows that our proposed dynamic adjustment strategy is able to provide different recommended actions for different agents. When three agents are in the same grid around Dongxiaoying, which is located outside the 5th ring road and has a relatively lower number of orders, the strategy guides the first two agents into the same grid but refers the third agent into a different grid because after guiding two agents into the same grid, the order matching probability in that grid drops quickly. The strategy also guides the first two agents into two different directions afterwards due to the competition. When three agents are in the same grid around Jinrongjie, which is located inside the 2nd ring road and has a relatively higher number of orders, the strategy guides all three agents into the same grid. Although the strategy refers the first agent into a different direction soon, the strategy guides the following two agents in almost the same direction.

As we have mentioned before, the historical number of orders in one grid is supposed to have an impact on the decrease of the order matching probability of the grid. For a grid with a higher average number of orders, like the grid located around Jinrongjie, the decrease in the order matching probability is supposed to be slower, thus it is able to accept more cruising agents while still maintains a relatively decent order matching probability. For a grid with a lower number of average number of orders, like the grid located in Dongxiaoying, the decrease in the order matching probability is supposed to be faster because with a small historical number of orders, it is not able to withstand too much competition among agents. In other words, for a grid with a small number of orders, the order matching probability is decreasing rapidly when some agents are being guided into the grid.

\subsection{Heterogeneity in reward functions} \label{sec:irl}

Although the determined coefficient $\alpha$ is applicable to the overall driver population at the aggregate level, it may vary from individual to individual. In other words, each driver may have a specific reward function which can be quite different from other drivers', resulting in some discrepancies in driving patterns and strategies. To examine various driving patterns of taxis drivers, especially the change in the driving patterns from the time without e-hailing to the time when e-hailing is widely adopted,  \cite{ma_qingyu_modeling_2019} carried out in-depth analysis using trajectories of taxi drivers in Shanghai, China, and unveiled that on average, taxi driving patterns which were previously concentrated in some central areas are now more spread out. Inspired by this, we examine the spread of a driver's trajectories using the radius of gyration $R_g$. Radius of gyration is defined as the standard deviation of the spatial distances between a driver's location and the centroid of the driver's visited locations, i.e., 
\begin{equation}
R_g = \sqrt{\dfrac{1}{n} \sum_{i = 1}^n r_i^2}
\end{equation}
where $r_i$ is the distance between the driver's $i^{th}$ location and the centroid of the driver's visited places, and $n$ is the total number of the driver's visited locations.

\begin{figure}[H]
	\centering
	\includegraphics[width=0.7\linewidth,height=\textheight,keepaspectratio]{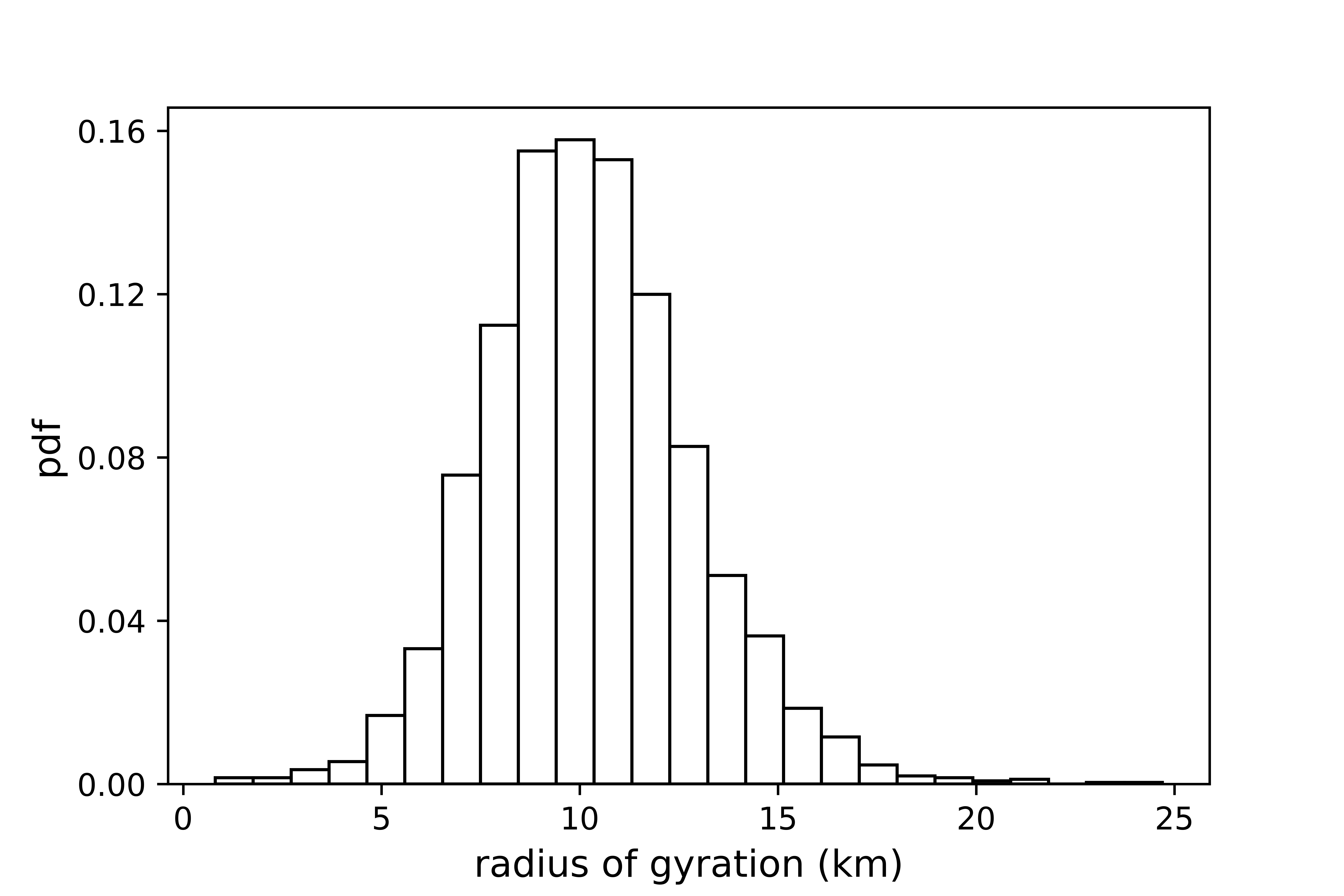}
	\centering 
	\caption{Distribution of radius of gyration}
	\label{fig:radius}
\end{figure}

Figure~(\ref{fig:radius}) plots the distribution of the radius of gyration across the e-hailing driver population. The radius of gyration of the majority (95\%) of e-hailing drivers is distributed within the range of $[5.5~\text{km}, 16~\text{km}]$. The e-hailing drivers on the left tail (2.5\%) of the distribution has a radius of gyration below $5.5~\text{km}$, and the drivers on the right tail has a radius of gyration above $16~\text{km}$. The substantial discrepancy in the radius of gyration across the driver population indicates that drivers exhibit different driving patterns, which may stem from different intrinsic reward functions. For example, the trajectory of a driver on the left tail with a small radius of gyration is more concentrated in a small region, indicating that the driver may have some incentives (such as being close to home or being more familiar with the region) to stay within the region.

%Since the goal of this work is not to characterize drivers with different driving patterns but to build an adaptive model which is able to capture the difference in driver's intrinsic reward function, 
To shed some light on the distinction among driving patterns of e-hailing drivers, we simply take one driver on the left tail with a small radius of gyration ($1.2~km$) and one driver on the right tail with a large radius of gyration ($17.8~km$) and employ the IRL technique to disclose some underlying information regarding the driver's intrinsic reward function. Here we emphasize that understanding drivers' intrinsic reward function can be very helpful from the perspective of the platform to appropriately assign e-hailing orders. For instance, for a driver who prefers to stay within a small region, the platform is supposed to assign relatively shorter e-hailing orders to this driver. Otherwise, the driver will have a lower utility and may even be unwilling to take the assignment since the driver cares more about staying within the region over a simple higher monetary return. Thus, appropriately assigning orders can help increase the drivers' response rate and therefore the utilization of vehicle resources, as well as the passenger satisfaction, which is beneficial for all players in the e-hailing market, including the platform, the drivers, and the passengers.

Without any knowledge of the driver's demographic information, we devise the third simple reward function $\phi_3(s, s')$ as the difference between the spatial distance between state $s$ and the centroid $c$ of the driver's visited locations and the spatial distance between state $s'$ and the centroid $c$, i.e.,
\begin{equation}
\phi_3(s, s') = distance(s, c) - distance(s',c)
\end{equation}      
where $distance(s,c)$ denotes the spatial straight line distance between $s$ and $c$. The rationale of this simple reward function can be explained as follows. For a driver with a small radius of gyration, the driver prefers to come back around the centroid after completing an order (otherwise the radius of gyration would be larger), indicating that coming back to the centroid may increase the driver's utility/intrinsic reward. $\phi_3$ exactly does this. The driver gets a positive reward when $distance(s',c) < distance(s,c)$ and a negative reward when $distance(s',c) > distance(s,c)$, meaning that it is beneficial for the driver to go back to the centroid. While for a driver with a large radius of gyration, the driver may not care about his/her distance to the centroid, and thus $\phi_3$ may not be important in the driver's intrinsic reward function. We apply the IRL technique to the observed policy of these two drivers with three simple reward functions, namely, $\phi_1$ (fare), $\phi_2$ (traveling distance), and $\phi_3$, and the derived coefficients are listed in Table~(\ref{tab:coeffs}). Again, we fix $\alpha_1 = 1$ under the assumption that the driver gets all the fare. %Note that the goal is to reveal the relative importance of each simple reward function, we thus restrict the summation of the coefficients to be 1, i.e., $\sum_{i = 1}^3 \alpha_i = 1$.}   

\begin{table}[H]
	\centering
	\caption{Coefficients}\label{tab:coeffs}
	\begin{tabular}{|p{4.5 cm}|p{3 cm}|p{3 cm}|p{3 cm}|}
		\hline
		& $\phi_1(s, s')$ & $\phi_2(s,s')$ & $\phi_3(s,s')$ \\
		\hline
		The driver with a small radius of gyration & $1.00$ & $0.21$ & $1.42$  \\
		\hline
		The driver with a large radius of gyration & $1.00$  & $0.37$ & $0.17$ \\
		\hline
		%The driver on the right end of the distribution & $0.67$  & $0.31$ & $0.02$ \\
		%\hline
	\end{tabular}
\end{table}	

The coefficient for $\phi_3$ is quite large ($1.42$) for the driver with a small radius of gyration and relatively small ($0.17$) for the driver with a large radius of gyration. This validates the effectiveness of the devised reward function $\phi_3$ in explaining the driver's intrinsic reward function when the driver's radius of gyration is small. When the driver's radius of gyration is large, the reward function $\phi_3$ does not contribute enough to the driver's underlying reward, meaning that the driver with a large radius of gyration does not gain more utility/reward by coming back to the centroid. We believe that there exist other types of simple reward functions which may be important in explaining a driver's underlying reward function when the driver has a large radius of gyration and will be left in future work.

\iffalse
\subsection{Model limitation}
Both the driver's rate of return and the utilization rate of the vehicle can be improved through the proposed modified MDP model. There are some extensions can be done to overcome some limitations: 
\begin{enumerate}
	\item Although a dynamic adjustment strategy for multiple agents has been devised and validated, the model does not fully incorporate the dynamic real-time multi-agent competition. In other words, since all parameters (i.e., probabilities) are predetermined, the model is not able to fully capture the competition among agents, resulting in potential overestimation issues. A multi-agent reinforcement learning approach can thus be adopted to consider the real-time competition and can yield a more realistic optimal policy for recommendation.
	\item The MDP model uses the aggregated historical data and ignores the temporal variation of the variables, such as $p_{order\_match}$, $p_{pickup}$, etc., within each time interval, which may cause bias in the estimation of variables. A Q-learning approach which uses the online info of the historical data can be a potential tool for tackling the problem. 
	\item When the grid size is small, the number of states in the MDP can be large, resulting in a higher requirement in the computation power. A hierarchical MDP model which reduces a big problem into several subproblems can be an efficient tool. For example, we can first divide the whole space into big zones and then divide each zone into finer grids.
\end{enumerate}
\fi

\section{Conclusion} \label{sec:conclusion}

Based on a large-scale real-world historical GPS traces, this paper investigated how to improve the income and rate of return of e-hailing drivers through a modified MDP approach. We proposed an MDP model which incorporates the following distinct features of drivers with e-hailing: (1) An e-hailing driver may receive a matched order before she drops off the previous passenger, thus there is no passenger seeking; (2) Different from traditional taxi that a driver has to see a passenger to find a match, e-hailing platforms very likely find a match even when the driver and the passenger are spatially far from each other. In other words, a driver's search process may end before a passenger is picked up.
 %Furthermore, in stead of deriving a deterministic policy, which will guide all drivers towards the same location, resulting in excess taxi supply at some areas and a localized competition, we employed a randomized policy to alleviate the competition. By adjusting the temperature parameter, one can obtain a near-deterministic policy from the derived randomized policy. 

We used 44,160 Didi drivers 3-day trajectories to train the model with a reward function uncovered by IRL. We then examined the optimal policy learned from our model and found that with e-hailing, the optimal policy suggests drivers to stay when they are in the city area, to move to some local areas with a high probability of receiving a request if they are in suburban areas, and to wait when they are at some places with a very high likelihood of receiving a request. To validate the effectiveness of the derived policy, a Monte Carlo simulation is conducted, and two metrics, namely the rate of return and utilization rate, are employed to compare the performance of the agent following the derived optimal policy with that of the agent following one baseline heuristic, namely, the local hotspot strategy. The comparison validates the effectiveness of the proposed model and shows that our model is able to achieve a 17.5\% improvement and a 7.5\% improvement over the local hotspot strategy in terms of the rate of return and the utilization rate, respectively. Also, the results show that under the guidance of the optimal policy, the agent is able to complete more order, decrease idling time, and find better orders. In addition, we disclose the reason why the agent not necessarily prefers a shorter order. 

In the modified MDP model, the order matching probability captures the supply-demand ratio by its definition, considering the fact that the number of drivers in this study is sufficiently large and thus the number of unmatched orders is assumed to be negligible. Results show that the optimal policy does not necessarily guide an e-hailing driver to a grid with a high demand. Instead, the optimal policy suggests a driver to a grid with a low supply-demand ratio, i.e., a high order matching probability. To accurately capture the competition among drivers, we have devised and calibrated a dynamic adjustment strategy of the order matching probability when there are multiple drivers need recommendations. Also, the heterogeneity in reward functions across the driver population has been investigated. The devised simple reward function $\phi_3$ is validated to explain the driving patterns of drivers with a relatively small radius of gyration and paves the way for future research on the underlying reward function of e-hailing drivers.

There are some extensions can be done to overcome some limitations of the model:
\begin{enumerate}
	\item Although a dynamic adjustment strategy for multiple agents has been devised and validated, the model does not fully incorporate the dynamic real-time multi-agent competition. In other words, since all parameters (i.e., probabilities) are predetermined, the model is not able to fully capture the competition among agents, resulting in potential overestimation issues. A multi-agent reinforcement learning approach can thus be adopted to consider the real-time competition and can yield a more realistic optimal policy for recommendation.
	%\item The MDP model uses the aggregated historical data and ignores the temporal variation of the variables, such as $p_{order\_match}$, $p_{pickup}$, etc., within each time interval, which may cause bias in the estimation of variables. A Q-learning approach which uses the online info of the historical data can be a potential tool for tackling the problem.
	\item When the grid size is small, the number of states in the MDP can be large, resulting in a higher requirement in the computation power. A hierarchical MDP model which reduces a big problem into several subproblems can be an efficient tool. For example, we can first divide the whole space into big zones and then divide each zone into finer grids. 
	\item A more driver-specific/personalized reward function can be investigated and is expected to incorporate more factors such as safety, stress, etc. Coupled with some prior knowledge, a more powerful IRL approach, such as the maximum entropy IRL  \citep{ziebart_maximum_2008}, could be an promising way to uncover more information of the underlying reward function that results in the demonstrated behavior. 
\end{enumerate}

\section*{Acknowledgments}

This research was funded by Didichuxing under the contract No. University of Michigan/DiDi 17-PAF07456. 
The authors would like to thank Mr. Partrick Guan, Mr. Hanyuan Shi, and Mr. Zhaobin Mo to prepare Didi data and run codes on Didi servers for us.  

\bibliographystyle{elsarticle-harv}
\bibliography{mdp}

\end{document}